%% file: paper.tex
\documentclass[review]{elsarticle}

\usepackage{booktabs}
\usepackage{multirow}
\usepackage{makecell}

\usepackage{graphicx}
\usepackage{subcaption}

\usepackage{tabularx}

\usepackage{multirow}

\usepackage{amsfonts}
\usepackage{bbm}

\usepackage{lineno,hyperref}
\modulolinenumbers[5]

\usepackage{changes}
\definechangesauthor[name={Mattia}, color=blue]{mattia}

\journal{Pervasive and Mobile Computing}

\begin{document}

\begin{frontmatter}

\title{MyDigitalFootprint: an extensive context dataset for pervasive computing applications at the edge}

\author[add1]{Mattia G. Campana\corref{cor1}}
\ead{m.campana@iit.cnr.it}
\author[add1]{Franca Delmastro}
\cortext[cor1]{Corresponding author}

\address[add1]{Institute for Informatics and Telematics of the National Research Council of Italy (IIT-CNR), Via Giuseppe Moruzzi, 1 56124 Pisa, Italy}

\begin{abstract}
The widespread diffusion of connected smart devices has greatly contributed to the rapid expansion and evolution of the Internet at its edge, where personal mobile devices follow the behavior of their human users and interact with other smart objects located in the surroundings. In such a scenario, the user context is represented by a large variety of information that can rapidly change, and the ability of personal mobile devices to locally process this data is fundamental to make the system able to quickly adapt its behavior to the current situation. This ability, in practice, can be represented by a single elaboration process integrated in the final user application, or by a middleware platform aimed at implementing different context processing and reasoning to support third-party applications.   
However, the lack of public datasets that take into account the complexity of the user context in the mobile environment strongly limits the advance of the research in this field.

In this paper, we present MyDigitalFootprint, a novel large-scale dataset composed of smartphone embedded sensors data, physical proximity information, and Online Social Networks interactions aimed at supporting multimodal context-recognition and social relationships modeling. The dataset includes two months of measurements and information collected from the personal mobile devices of 31 volunteer users, in their natural environment, without limiting their usual behavior.Existing public datasets generally consist of a limited set of context data, aimed at optimizing specific application domains (human activity recognition is the most common example). On the contrary, our dataset contains a comprehensive set of information describing the user context in the mobile environment.
In order to demonstrate the efficacy of the proposed dataset, we present three context-aware applications based on different machine learning tasks: (i) a social link prediction algorithm based on physical proximity data, (ii) the recognition of daily-life activities based on smartphone-embedded sensors data, and (iii) a pervasive context-aware recommender system.
To the best of our knowledge, this is the first large-scale dataset containing such heterogeneity of information, representing an invaluable source of data to validate new research in mobile and edge computing.
\end{abstract}


\end{frontmatter}


\section{Introduction}

Nowadays we are living surrounded by a plethora of electronic devices that are equipped with a continuously increasing amount of both computational and networking capabilities.
In particular, the computational resources of personal mobile devices (e.g., smartphones, tablets, and wearables) are comparable or sometimes exceed those of desktop computers of the past years. In addition, their multiple communication interfaces guarantee both a continuous Internet access and proximity-based communications opportunities.
These conditions, along with the widespread penetration of Internet of Things devices (IoT) embedded in physical objects, contribute to a rapid expansion and evolution of the Internet at its edge, and leads to the raise of new paradigms for the future Internet~\cite{7273478}.
For example, based on the observation that the edge of the Internet is mainly composed of personal mobile devices that follow the behaviour of their human users, the \emph{Internet of People} (\emph{IoP}) paradigm calls for a radical change of the Internet, where personal mobile devices are no more considered as simple clients~\cite{CONTI201851}.
Indeed, in this envisioned scenario, they represent active elements of the new Internet that are able to forward and disseminate data within the network by exploiting their wireless equipment and self-organizing networks~\cite{7123563, basagni2013mobile}, extending the users connectivity opportunities, including direct communications with other users and devices in proximity.

These aspects are paving the way towards new types of computing models, shifting most of the tasks from centralized architectures (e.g., remote servers and cloud-based computing) to distributed solutions, where the data available at the edge is directly processed by mobile devices~\cite{7807196}.
This paradigm shift creates new opportunities for the creation of novel pervasive mobile applications (e.g., data dissemination algorithms~\cite{7898395}, forwarding protocols~\cite{5677535}, and personalized services~\cite{ARNABOLDI20173, 10.1145/3298689.3347067}), which benefit from low-latency direct communications and the sensing capabilities of modern mobile devices.
Specifically, the great variety of sensors embedded in the personal mobile devices of the users provides essential information to recognize the context and the situation in which the user is involved, making context-awareness a real feature of new pervasive computing applications. 
A basic example can be the automatic configuration of the device based on the user's current activity, while more complex context information must be analyzed to optimize networking and forwarding protocols to disseminate content according to users' interests and social relationships.

Processing user data directly on the local device provides two main advantages. Firstly, it allows both the device and the applications to quickly adapt their behavior according to the changes in the user context.
Even though the context recognition task can be performed on remote servers, data transmission delay may make the computation useless with respect to the service optimization, since the user could have changed her context in the meanwhile.
Secondly, user privacy can considerably benefit from the use of such a decentralized approach~\cite{7469991}.
Indeed, traditional client-server solutions may demotivate privacy-aware users to use context-aware services since a third-party entity will be in charge of storing and processing users data, and additional mechanisms must be employed to safeguard the user's privacy~\cite{6187862}.
On the contrary, shifting the computation from remote servers to the source of the context data (i.e., the user's mobile device) allows the system to preserve the user's privacy, avoiding the need of trusting external systems.
In the last few years, such advantages have been found in different application domains, including IoT systems~\cite{7498684}, healthcare~\cite{9124720}, smart cities~\cite{8675180}, and, more in general, context-aware and mobile applications~\cite{8016573}, highlighting the need of a paradigm shift.

However, to validate and evaluate the effectiveness of context-awareness in pervasive computing applications, the availability of context data collected in the wild becomes an essential requirement.
Most of the public datasets available in the literature focus on few sensors data for inferring the user's context.
For instance, keeping into account only the accelerometer and gyroscope data is a common practice for recognizing simple human activities and transportation modalities~\cite{6838194, 10.14778/2733004.2733015}, while the GPS coordinates and the list of Wi-Fi access points are commonly used to infer the current location of the user and her social interactions~\cite{10.1145/2370216.2370288, DEDOMENICO2013798}.
On the other hand, the user's context in a mobile setting represents a more abstract concept, which requires the combination of heterogeneous sources of data that characterize not just the user's activities, but also her behavior, her social interactions with other people, her daily-life situations and the surrounding environment.
For this reason, in recent years researchers started to collect a wider range of smartphones and wearable sensors data.
Several research studies in the area of context-recognition and human behavior modeling base their results on experiments performed in controlled environments (e.g., a research laboratory), with researchers instructing subjects to perform scripted tasks, generally using the same device~\cite{app7101101, anguita2013public, 7134104}.
However, lab results typically diverge from those obtained in real-world experiments, in which users may have different ways of performing the same activity, and devices are equipped with different types of sensors~\cite{Kerr2016}.
In addition, according to~\cite{8090454}, to better represent the complexity of the real-world, context data should be collected in natural and realistic settings, satisfying the following in-the-wild conditions: (i) to represent the variety of available devices, the subjects should not be forced to use a foreign device, but they should use their smartphones; (ii) to address the variability in device wearing and placement, no restrictions on device usage should be defined; and (iii) the recorded data should represent the users' natural behavior, thus they should not be instructed on how to perform the activities.

In this work, we present \emph{MyDigitalFootprint} (\emph{MDF}), a novel large-scale dataset that we collected from the personal smartphones of 31 volunteer users within a period of 2 months.
Following the in-the-wild data collection protocol, we installed on the volunteers' devices an Android sensing application that monitored a wide range of heterogeneous smartphone sensors in the background, without interfering with the user's natural behavior.
More specifically, the application continuously collected data from both physical and virtual sensors that can be used to characterize the different aspects of the user context in a mobile setting, including daily-life situations and social interactions with other people.
Physical sensors refer to phone-embedded hardware (e.g., accelerometer and gyroscope) aimed at describing simple human activities (e.g., the user's gait), while virtual sensors represent data sources that characterize the device status, the surrounding environment, and the interactions between the user and her device.
In order to collect data that actually represents users’ daily-life situations, we defined no constraints related to the user behavior or the interactions with her device during the experiment.
On the contrary, we encouraged the volunteers to use their smartphones as usual, without worrying about the positions of the device (e.g., trousers pockets, or hand) or the activities they usually perform during the day.

Nonetheless, daily activities represent only part of the user's context.
In this paper, we argue that taking into account information related to the user's social relationships is fundamental to model the overall context in mobile environments.
In fact, inferring the diverse social ties in the interpersonal network of a subject can lead to better recognition of her context, helping to precisely discriminate among different daily-life situations~\cite{Zhang2016}.
For this reason, during the sensing experiment, we also collected essential information to model the users' social contexts: proximity data, and activities performed by the volunteers on Online Social Networks (OSN) platforms.
Specifically, we collected proximity information through the use of wireless communication interfaces (i.e., WiFi-Direct and Bluetooth scans), which can be used to infer face-to-face interactions among the people.
As far as OSN is concerned, they represent a rich source of data to characterize both the user's preferences and her social relationships with other people in the virtual world, such as contents shared by the users, their reactions (e.g., likes), comments, and the list of followed public profiles.
To the best of our knowledge, MDF is the first dataset presented in the literature that contains such heterogeneous data, aimed at modeling all the aspects of the user's context, combining information generated in both the physical and the cyber worlds.

In order to demonstrate the utility and the efficacy of MDF, we propose three different pervasive mobile applications based on machine learning methods.
Firstly, we propose a social interaction prediction algorithm based on physical proximity data collected by mobile devices.
This represents a key feature to design effective data dissemination and forwarding algorithms for the edge of the Internet.
Then, we define a context-recognition system that identifies the situation in which the user is currently involved based on multimodal and high-dimensional sensor features.
Finally, a pervasive context-aware recommender system for mobile devices is proposed. 
Implementing such a system directly on mobile devices allows to evaluate and automatically filter the contents discovered at the edge, providing personalized recommendations to the user, based on her current context and needs.

The entire dataset (i.e., 2.64 GB of data) is publicly available online~\footnote{\url{https://github.com/contextkit/MyDigitalFootprint}}, and researchers are encouraged to use it to develop and compare methods and algorithms in different research fields.
Along with the anonymized dataset, researchers will also find the processed data to replicate the proposed pervasive applications. 

As a summary, this work provides several contributions to the research areas of human behavior modeling and pervasive mobile computing, as follows:
\begin{itemize}
    \item a large-scale public dataset collected from commercial smartphones in the users' natural environment, without setting any constraint to the users' behavior;
    
    \item the dataset is characterized by the unique feature of combining heterogeneous smartphone sensors data with information derived from Online Social Network platforms;
    
    \item the use of machine learning algorithms to implement three different pervasive computing applications evaluated by using the MDF dataset: (i) prediction of physical interactions among the users, (ii) automatic recognition of the user's context, and (iii) context-aware recommendations in mobile environments;
    
    \item three different datasets extracted from MDF that has been already preprocessed to reproduce the proposed machine learning applications.
\end{itemize}

The remainder of the paper is organized as follows.
In Section~\ref{sec:related_work} we provide an extensive review of the existing datasets collected from mobile devices in the wild.
Section~\ref{sec:data_collection} describes the details of the data collection campaign and the experimental protocol used to collect the MDF dataset.
In Section~\ref{sec:data_analysis} we perform both a quantitative and qualitative analysis of the main data collected from the volunteers' smartphones.
Then, we rely on the presented dataset to design and evaluate three potential edge computing applications in Section~\ref{sec:applications}.
Finally, in Section~\ref{sec:conclusion}, we draw our conclusions and present some directions for future work.

\section{Related Work}
\label{sec:related_work}

Most of the public datasets related to human behavior modeling has been collected in controlled environments and focuses on a limited number of smartphone-embedded sensors aimed at recognizing a predefined set of human activities and events (e.g., user's gait or fall detection)~\cite{6838194, s150102059}.
Instead, the number of available datasets for the identification of a wider user's contexts, especially in outdoor environments, is limited.
Specifically, after an extensive search in the literature, we identified only the following datasets based on smartphone-embedded sensors: 
 \emph{ExtraSensory}~\cite{8090454}, \emph{ULSTER HAR}~\cite{9060182}, \emph{Sherlock}~\cite{10.1145/2996758.2996764}, and \emph{ContextLabeler}~\cite{10.1145/3267305.3274178}.
ExtraSensory is composed of heterogeneous data derived from the personal smartphones of 60 subjects (34 iPhone and 26 Android devices).
In addition, 56 volunteers agreed to use also a wearable device (i.e., a Pebble smartwatch) to capture additional user-device interactions and to increase the sensing capabilities of the entire experiment.
A mobile application installed on the users' smartphones collected both sensor measurements and ground truth context labels selected by the users.
Every minute, the application recorded a 20-second window of sensor measurements from both the smartphone and the smartwatch, asking the subject to specify her current activity.
In addition, the application raised periodic notifications, reminding the user to provide missing labels.
Users were able to describe their contexts by choosing the labels among those in the following categories: (i) a main activity label among 7 possible choices describing movement and posture of the user (e.g., \texttt{lying down}, \texttt{sitting}, or \texttt{standing in place}), and (ii) one or more secondary activity labels describing higher level contexts including sports (e.g. \texttt{playing basketball}, \texttt{at the gym}), transportation modalities (e.g. drive - \texttt{I’m the driver}, \texttt{on the bus}), basic needs (e.g. \texttt{sleeping} and \texttt{eating}), social environment (e.g. \texttt{with family}, \texttt{with co-workers}), and location (e.g. \texttt{at home}, \texttt{at work}).

The volunteers have been engaged for approximately one week with the request to keep the app running in the background as much as possible, and report as many labels as possible for each collected data sample.
The resulting dataset contains a total of 300 minutes of measurements, where each data sample is composed of 228 features extracted from both physical and virtual sensors, including accelerometer, gyroscope, user location, and phone state (e.g., app status, battery state, and Wi-Fi availability).
Thanks to the variety of the collected data, ExtraSensory has been used in several research works for the evaluation of human activity recognition systems~\cite{10.1145/3277593.3277617, s19071716}, personalization of machine learning models~\cite{8771081}, and behavioral assessments for e-health applications~\cite{8928614}.

Even though ExtraSensory contains data generated by a considerable amount of users, it is affected by two main drawbacks.
Firstly, since users were able to deactivate their smartphone sensors during the data collection to save battery or for privacy reasons, the resulting dataset is very sparse (i.e., 15\% of sparsity index).
Each data sample has at least a missing feature, and this makes the application of machine learning algorithms a challenging task.
Secondly, the mobile application did not collect data related to important sensors like, for example, the Wi-Fi Direct and Bluetooth, which are necessary to characterize the physical proximity among the users and devices, thus inferring their social interactions in the real world.

ULSTER HAR dataset has been collected by following the same protocol of ExtraSensory: 10 volunteer users acting in free-living and unconstrained conditions for a period of 6 weeks.
In this case, a mobile application has been used to monitor a small amount of physical smartphone-embedded sensors (i.e., accelerometer, gyroscope, GPS, and light sensor), while more abstract information has been acquired from a wearable activity tracker (i.e., a Nokia Pulse O2 smartband), including the number of steps, walking distance, burnt calories and sleep quality.
Users labeled the collected data samples with simple daily-life activities, such as, \texttt{sitting}, \texttt{standing}, \texttt{running}, \texttt{cycling}, \texttt{walking upstairs/downstairs}, and the use of some transportation modalities (e.g., \texttt{in vehicle} and \texttt{cycling}).
The authors used the collected data to define and evaluate a deep learning model for the recognition of simple human daily activities.
Specifically, they cross-validated a neural network trained on Extrasensory and tested its performance on the data samples contained in their ULSTER HAR dataset.
The obtained results highlighted high cross-subject variability when testing the model on new subjects, with a balanced accuracy varying between 53.33\% and 90.01\%, and an average accuracy of 71.73\%.

The Sherlock dataset has been collected from 50 users for cybersecurity research.
While a mobile application, called Moriarty, simulated malicious behaviors on the volunteers' devices, a second application, called SherLock, collected a great variety of context data generated by a heterogeneous set of sensors, including physical sensors (e.g., gyroscope and barometer), Bluetooth and Wi-Fi scans, mobile applications usage statistics, call and SMS logs, users' locations, and phone status (e.g., battery and display configurations).
With its 10 billion data recorded in 2 years, Sherlock represents one of the biggest and longest-lasting data collection experiment related to smartphone usage data.
It has been used to evaluate different research proposals, including energy-aware resource allocation at the edge~\cite{8897679, 8567678}; biometric classification and user identification, based on smartphone usage data~\cite{8669790, s18041219, s19112466}; and prediction of users' locations in mobile environments~\cite{8641122}.
However, the main limitation of the sensing experiment is related to the homogeneity of the mobile devices.
In fact, all the participants have been equipped with the same smartphone model (i.e., a Samsung Galaxy S5), and were asked to use it as their main phone.
According to~\cite{10.1145/2809695.2809718}, diversity in smartphones and sensor hardware highly affects the measurements.
For this reason, in our sensing experiments, we chose to use the volunteers' personal smartphones as context data sources, in order to take into account a great variety of models and sensing equipment.

As the last example, we also refer to our preliminary experience in the collection of smartphones sensors data in the wild, which generated the ContextLabeler dataset presented in~\cite{10.1145/3267305.3274178}. 
The dataset contains a total of 36354 data samples, annotated by 3 voluntary users with 8 different labels describing their daily-life activities and situations: \texttt{Home}, \texttt{Working}, \texttt{Sleeping}, \texttt{Restaurant}, \texttt{Break}, \texttt{Lunch}, \texttt{Free Time}, and \texttt{Shopping}.
Each sample is represented by a high-dimensional vector composed of 1332 features extracted from a wide set of both physical and virtual sensors, describing different characteristics of the user's context in a mobile environment: the status of the mobile device (e.g., display and battery), user-smartphone interactions (e.g., running applications), the surrounding environment, and other devices in proximity discovered through the use of both Wi-Fi Direct and Bluetooth technologies.
During a period of two weeks, the volunteers used a specific mobile application installed on their smartphones that allowed them to freely specify their daily activities, while it continuously collected sensor data unobtrusively, monitoring all the data sources in the background.
In our previous work~\cite{10.1145/3267305.3274178}, we used ContextLabeler to validate a lightweight approach to model the user context able to efficiently perform the entire context-recognition process on the user mobile device.
Specifically, by using dimensionality reduction algorithms (e.g., PCA and Autoencoder) we have been able to speed-up the reasoning process by ten times compared to classifiers that use raw features while keeping the classification accuracy loss less than 3\%.

The MDF dataset presented in this paper significantly differs from the other research works with regards to the variety of the collected data, especially related to users' social interactions, both in the physical and the cyber world.
Social data includes information related to face-to-face interactions among the users and their activities on OSN platforms, which represents an invaluable source of information to characterize both the users' social relationships and their preferences.
To the best of our knowledge, MDF is the first large-scale dataset proposed in the literature that combines heterogeneous smartphone sensors data collected in the wild, with information related to both proximity data among the users and their interactions on OSN.
Therefore, the proposed dataset represents a precious source of information that takes into account all the aspects of the user's context, creating a bridge between the physical and the virtual worlds.

\section{The MyDigitalFootprint dataset}
\label{sec:data_collection}

To obtain a dataset that highly represents the real mobile environment, we designed a data collection campaign following the in-the-wild protocol.
To this aim, we enrolled a total of 31 volunteer users (6 females and 25 males), where the majority of them (i.e., 24 people) were students between age 14 and age 17 involved in a training period at research institutes, coming from high-schools of three different cities located in the Tuscany region of Italy (i.e., Pisa, Pontedera, and Livorno). 
The other 7 volunteers were Ph.D. students and researchers working at the National Research Council of Italy located in Pisa, Italy.
According to the EU GDPR, to participate in our sensing experiment, all the users signed an informed consent that includes all the policies adopted for personal data storage, management, and analysis, including the publication of the anonymized dataset.

Users manually installed on their mobile devices an Android application that we especially designed to monitor a wide range of smartphone-embedded sensors unobtrusively, collecting all the data in the background to not interfering with the natural user-device interaction.
Moreover, we encouraged the volunteers to naturally use their smartphone, without worrying about the location or position of the device, or instructing them to perform some predefined activities.

\begin{figure}[t]
    \centering
    \includegraphics[width=\textwidth]{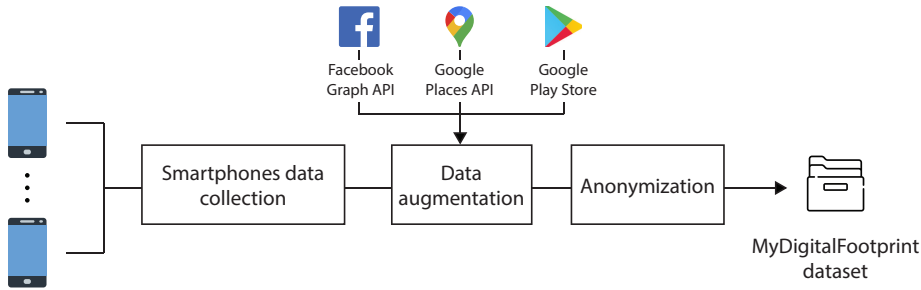}
    \caption{Overview of the dataset creation process.}
    \label{fig:dataset_creation}
\end{figure}

Figure~\ref{fig:dataset_creation} shows the high-level process we followed to produce the MDF dataset.
Firstly, we  collected all the sensors data generated by the users' smartphones over the 2 months of sensing experiment.
Second, we extended the smartphones data with labels describing the users' context, based on their location, and additional information downloaded from the Internet.
More precisely, we downloaded different types of information related to the users' activities on OSN platforms, including their shared contents, interactions with other people, and their interests.
Then, we enriched the location data with the information related to the most probable places visited by the users.
Finally, we extended the information related to the mobile applications used by the volunteers during the data collection, downloading their reference categories from the Google Play Store~\footnote{\url{https://play.google.com/}}.
As a final step, we anonymized the collected data to preserve the user's privacy.
More specifically, we have replaced all the personal information with fictitious unique identifiers, including phone numbers, and OSN profile IDs.

\subsection{Data collection}

The data collection campaign officially started on March 27~\textsuperscript{th}, 2018, when we met the volunteers at the CNR research center to introduce them to the details of the sensing experiment and to install on their smartphones the data collector application.
During the application setup, we asked the participants to accept the informed consent in which the collected data, the target of the study, and the data management process were extensively explained.
In addition, the users must accept all the necessary Android permissions required to access the smartphone sensor data, including the authorization to interact with the selected third-party applications. Specifically, we asked the users to perform the log-in procedure through their Facebook account, in order to allow the application to collect their Facebook access tokens required to download the OSN data through the official API.
Figure~\ref{fig:android_app} shows a screenshot of application User Interface (UI).
The main purpose of the UI was to show a set of descriptive statistics about the user activities on the OSN and the collected context data.
Please note that, even though our application has been designed to collect information from different OSNs, the MDF dataset contains only data downloaded from Facebook, which was the OSN platform most used by our volunteers.

\input{fig_mdf_app}

To avoid affecting the user’s behavior, the mobile application collects all the sensors information in the background through the use of the sensing framework ContextKit (CK)~\cite{10.1145/3267305.3274178}.
CK has been designed to continuously monitor the status of a broad set of mobile sensors, including both physical and virtual sensors, that can be used to model the complexity of the user's context in mobile settings, including daily life activities, preferences, and physical interactions with other people (i.e., face-to-face contacts)~\footnote{The CK framework is an open-source project, and it can be downloaded from \url{https://contextkit.github.io}}.

According to the information that can be extracted from the sensors data, we can classify the sensors monitored by CK in the following 6 categories:

\paragraph{User activity} The user activity can be modeled by taking into account different information. Specifically:

\begin{description}
    \item[Location] Information related to the geographical location of the user, including latitude, longitude, location accuracy, and bearing (i.e., the horizontal direction of travel).
    
    \item[User's gait] The user's activity detected by the Android Activity Recognition system~\footnote{\url{https://developers.google.com/location-context/activity-recognition}}. Those activities include both human gait (e.g., \texttt{running} and \texttt{walking}), and transportation modalities (e.g., \texttt{in vehicle} and \texttt{riding a bike}).
    
    \item[Calendar] Events or appointments defined by the user in the Calendar app.
    
    \item[Multimedia] List of multimedia type (i.e., photo or video) generated by the user by using the smartphone camera.
    
     \item[Running apps] List of running applications (i.e., their package name).
\end{description}

\paragraph{Smartphone status} The current status of the smartphone in terms of audio system settings, battery information, display status, and physical sensors data can describe the situation in which the user is currently involved. For example, during a business meeting, the ringtone might be muted, and the display turned off.  

\begin{description}
    \item[Audio] Information related to the smartphone audio system configuration, including ringer modality (e.g., audible or silent), notification volume, and whether the speaker is switched on or off.
    \item[Battery] Information related to the smartphone battery status, including charging level and modality (e.g., plugged to a USB port or an AC charger).
    \item[Display] Status of the smartphone display (e.g., on, off, or in low-power mode) and its orientation.
    \item[Physical sensors Data] including environment sensors (e.g., ambient temperature and light), motion sensors (e.g., accelerometer and gyroscope), and position sensors (e.g., rotation and proximity). For each sensor, CK collects 200 samples and calculates different descriptive statics, such as percentiles, skewness, kurtosis, and variance.
\end{description}

\paragraph{Cellular network} In this category we can group information related to the cellular network, including visible cellular antennas and phone calls.
While the former is useful to characterize the user's location, the latter can be used to model her social relationships with other people.

\begin{description}
    \item[Calls] List of incoming and outgoing phone calls.
    \item[Network cells] List of visible mobile network cells. For each cell CK keeps track of the technology type (e.g, GSM or LTE), the cell ID (if available), and the signal strength.
\end{description}

\paragraph{Wireless interfaces} Wireless communication interfaces (e.g., Wi-Fi and Bluetooth) can be useful to model both the user’s social interactions (e.g., other users in proximity) and the situation in which she is involved.

\begin{description}
    \item[Bluetooth connections] Information related to the Bluetooth devices connected to the user's smartphone, including their physical address, and the device type identified by their Bluetooth class IDs.
    \item[Bluetooth scans] List of Bluetooth devices in proximity.
    \item[Wi-Fi P2P] List of Wi-Fi Direct devices in proximity.
    \item[Wi-Fi] list of Wi-Fi Access Points in proximity.
\end{description}

\paragraph{User interests} In a mobile setting, the user's tastes can be modeled by analyzing the type of application she commonly uses during the day.

\begin{description}
    \item[App usage] Statistics related to the usage of installed applications, including last time usage and total time spent in the foreground.
    \item[Installed apps] List of installed applications.
\end{description}

\paragraph{External information} Information not strictly related to the mobile device can be used to refine the user’s context. For example, the user’s activity and location might depend on the current weather conditions.

\begin{description}
    \item[Weather] Information related to the weather conditions obtained by using the OpenWeather API service~\footnote{\url{https://openweathermap.org/api}}, including weather status (e.g., raining or cloudy), temperature, humidity, and wind speed.
\end{description}

Table~\ref{table:sensors} shows the sampling rate that we used for each sensor during the data collection.
When the data collector application is running, CK squires new data samples in the range of [1-5] minutes for most of the sensors, while it updates the applications' usage statistics and the weather conditions every hour.
Moreover, the \emph{Bluetooth Connections}, \emph{Phone Calls}, and \emph{Multimedia} sensors react to specific events: when the user connects or disconnects a Bluetooth device to her smartphone; when she receives or makes a phone call; or when she takes a photo or records a video; CK records such information to specific log files.

In order to simplify the data collection in large-scale experiments, the mobile application leverages a specific feature of CK that periodically sends the collected data to a specified back-end server. Specifically, once the collected data reaches a predefined dimension (i.e., 2 MB in our case), CK compresses the log files in a single zip archive and sends it to the remote endpoint as soon as a Wi-Fi (and stable) connection is available.

\subsection{Data augmentation}
\label{sec:data_augmentation}

After two months of data collection, on May 28th, 2018, we met the volunteers for the second time at the CNR center to officially close the sensing experiment and to uninstall the data collector application from their smartphones.
Then, as a post-processing step, we enriched the information contained in the collected dataset by including the following information: (i) the users' activities and interactions on OSN, (ii) semantic information related to the most probable venues visited by the users, (iii) the type of mobile applications used during the experiment, and (iii) context labels describing the situation in which the users were involved, based on their location data.

OSN platforms are invaluable sources of data that can be used to model both the user's preferences and social relationships~\cite{6425617}.
Among the different OSNs, Facebook is one of the most used by the participants because it allows them to access a great variety of content.
On Facebook, each user can access a web page called timeline, where she can read the status updates (\emph{posts}) shared by her friends and interact with them.
Moreover, a user can follow the status updates of different public pages, specialized on different topics (e.g., movies, video games, and TV shows).
For each user, we used the Facebook Graph API~\footnote{\url{https://developers.facebook.com/docs/graph-api}} to download the following information from her OSN profile:

\begin{description}
    \item[Friends] List of volunteers who are also friends on Facebook
    \item[Shared contents] The IDs of the contents shared by the user, including photos, videos, and posts.
    \item[Interactions] For each shared content, we also downloaded the interactions with other users represented by comments and reactions (e.g., likes).
    \item[Public pages] List of public pages liked (and followed) by the user.
\end{description}

Regarding the user's locations, we have used the Google Places API~\footnote{\url{https://developers.google.com/places/web-service/intro}} to download the information related to the most probable venues visited by the users during the sensing experiment.
More specifically, we have collected the type of the places (e.g., \texttt{restaurant}, \texttt{grocery store} or \texttt{pharmacy}) closest to the locations indicated by the GPS coordinates contained in our dataset.
This information can be used to refine not just the context in which the user was involved during her daily life, but also to characterize her habits and preferences.

Then, we additionally characterized the information collected about the mobile applications used by our volunteers.
On the Google Play Store, each mobile application belongs to a specific category that indicates its main functionality.
For instance, the mobile client of Facebook belongs to the \texttt{SOCIAL} category, while the popular WhatsApp Messenger relates to \texttt{COMMUNICATION} application group.
As we discuss in Section~\ref{sec:applications}, this type of information is extremely relevant to model the user's context and her preferences.

\input{fig_map.tex}

Finally, we retrospectively labeled the context data samples, based on the information related to the users' locations.
Figure~\ref{fig:map_locations} shows the heatmap of the locations visited by the users.
The red square represents the geographical area where the users live, which is shown in detail in Figure~\ref{fig:loc_exp_area}, highlighting the three main cities in the area: Pisa, Pontedera, and Livorno.
According to the user's location, we defined the following context labels: \texttt{Home}, \texttt{School}, \texttt{Workplace}, \texttt{CNR meeting} (i.e., when we met the volunteers for the data collection experiment), \texttt{Free Time}, and \texttt{Holiday}.
The locations in which the user context corresponds to \texttt{School}, \texttt{Workplace}, \texttt{CNR meeting} are known; respectively, they refer to the following places: the three schools considered in the sensing experiment, the CNR research center located in Pisa, and a specific area of the CNR used for the meetings with the volunteer users.
For identifying the users' \texttt{Home} locations, we clustered their GPS data generated during the night, and then we selected the cluster with the highest number of data samples.
Finally, the information labeled with \texttt{Free Time} and \texttt{Holiday} corresponds to the remaining GPS traces that have been generated respectively inside and outside the geographical bounding box shown in Figure~\ref{fig:map_locations}.

\section{Data analysis}
\label{sec:data_analysis}

\begin{figure}[t]
    \centering
    \includegraphics[width=.8\textwidth]{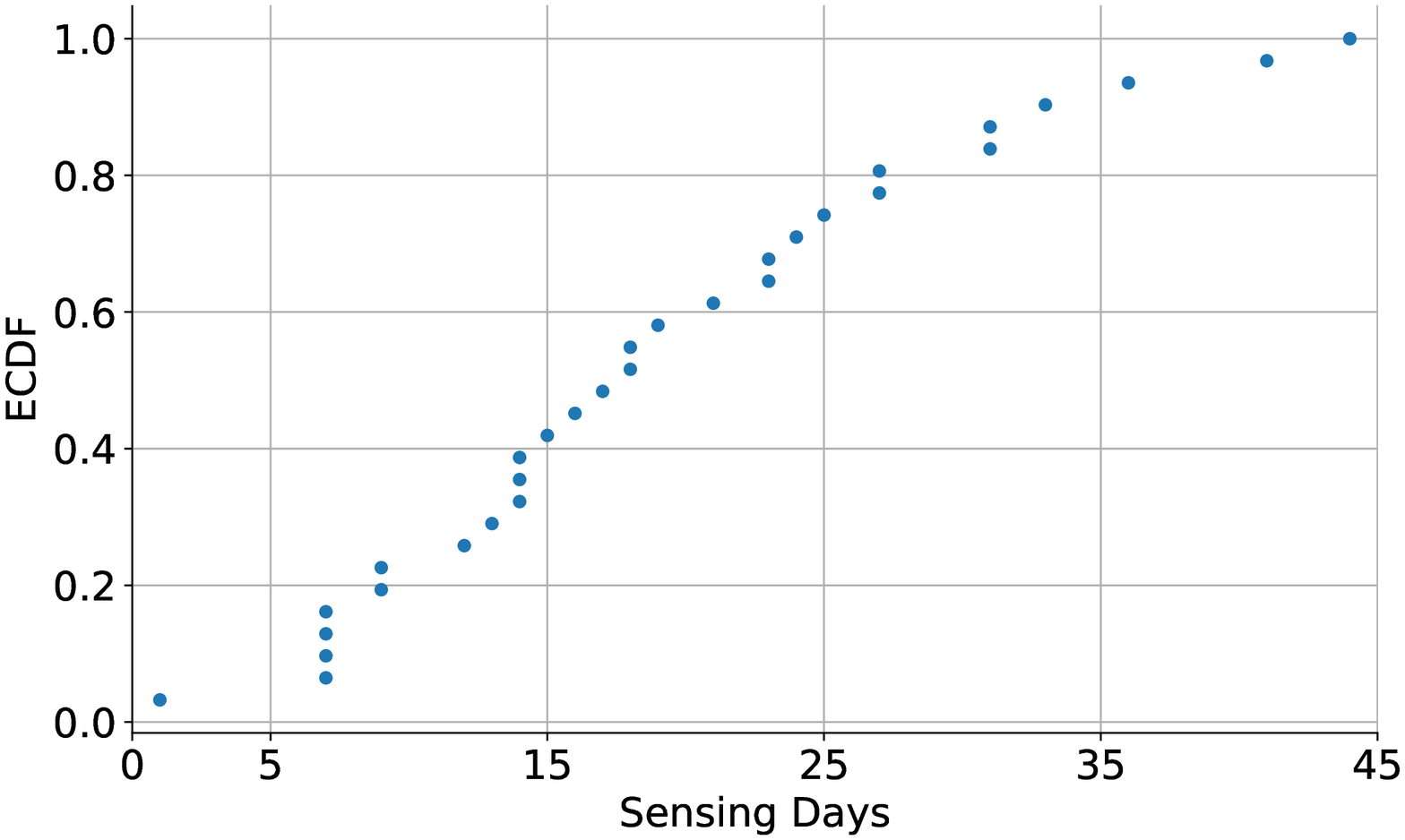}
    \caption{Distribution of the number of days collected for each user.}
    \label{fig:sensing_days}
\end{figure}

The presented dataset has been collected through the use of an Android application installed on the volunteers' devices to monitor a wide range of physical and virtual sensors.
Even though our application ran in an unobtrusively way (i.e., in the background) to collect the data without interfering with the user behavior, users were able to deactivate the data collection for privacy reasons.
Figure~\ref{fig:sensing_days} shows the distribution of the total number of days in which our application has been used by each volunteer.
As we can note, 12 volunteers (about 40\%) used our application for less than 15 days, but the great majority of the users have kept the application active for a period between two weeks and 45 days.

In this section, we provide both a quantitative and quality analysis of the main information collected in the MDF dataset.
Even though a more detailed characterization of the data is possible, we mainly focus on the correlation between the collected information and the users' context, which represents one of the main objectives of this work.

\subsection{Data annotation}
\label{sec:data_analysis_annotation}

\input{fig_da_labels}

As we described in Section~\ref{sec:data_augmentation}, at the end of the data collection campaign, we annotated the context information based on the users' locations.
Figure~\ref{fig:labels_weekdays} and Figure~\ref{fig:labels_weekend} show the distributions of the context labels during the weekdays and weekends, respectively.
We can note that the \emph{School} label distribution follows the typical time table of Italian schools: the majority of the data samples have been generated during weekdays, within 7 a.m. and 1 p.m., while few of them refer to the afternoon (between 1 p.m. and 5 p.m.) because of extra lessons or sports activities.
In addition, some students attend lessons also on Saturday morning.
Ph.D. students and researcher at the CNR do not have a strict working time table.
This is the reason why the distribution of \texttt{Workplace} label covers a wide range of hours during the week, starting from 8 a.m. until 9 p.m., where the majority of the data samples has been collected during typical working time (i.e., between 10 a.m. and 6 p.m.).

Regarding the \texttt{Home}, \texttt{Free Time} and \texttt{Holiday} labels, we can note a clear distinction between the distributions shown in the two figures.
During the week our users spend more time at home, especially during the evening and night, where the percentage of data samples annotated with \texttt{Home} reaches 80\% as the maximum value.
On the other hand, during the weekend, the \texttt{Home} distribution slightly exceeds 60\%, while the rest of the data samples is annotated with \texttt{Free Time} and \texttt{Holiday}.

Finally, samples annotated with \emph{CNR meeting} refer to the meetings we organized with the students at the beginning and at the end of the data collection campaign.
For this reason, the total amount of data samples annotated with this label is the 17\% of the total dataset only, and its distribution ranges from 9 a.m. until 5 p.m. on weekdays.

\subsection{Display status}

\begin{figure}[t]
    \centering
    \includegraphics[width=\textwidth]{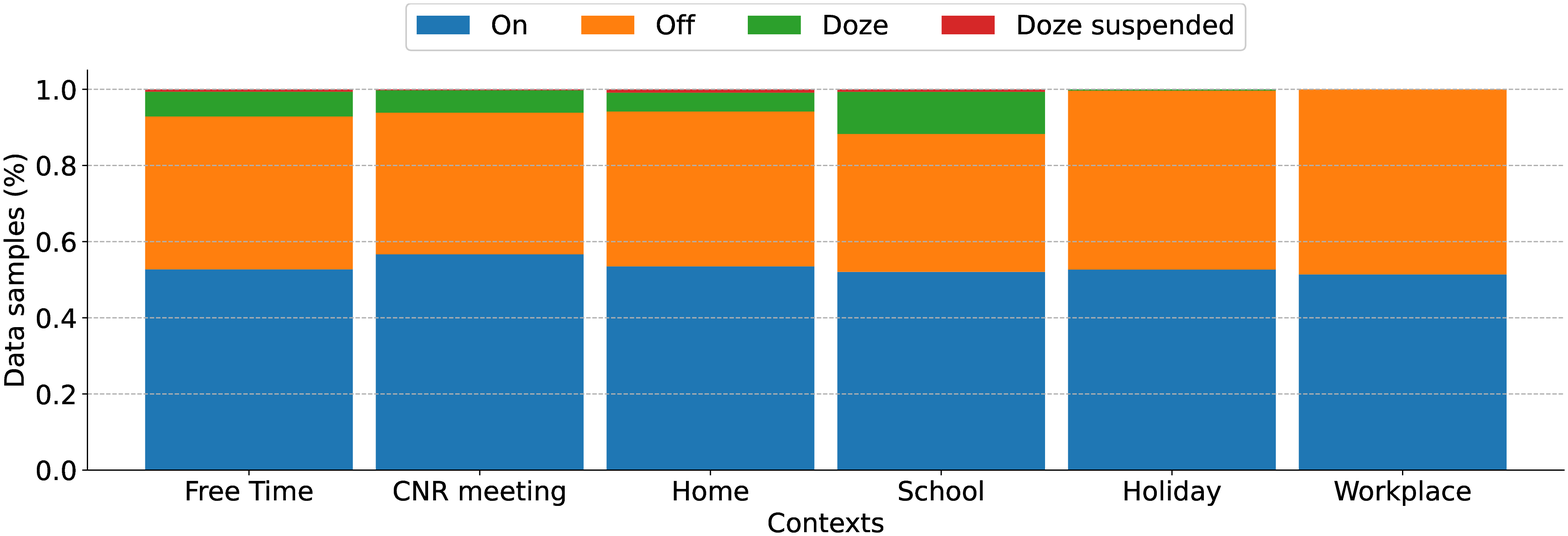}
    \caption{Display status in the different contexts.}
    \label{fig:display_status}
\end{figure}

Figure~\ref{fig:display_status} shows the number of display data samples for each users' context, highlighting the different status in which the display of an Android device could be: \emph{On}, it is active, \emph{Off}, it has been turned off, \emph{Doze}, the display is in low-power mode and it displays only system-provided content, and \emph{Doze suspended}, it is in low-power mode and the CPU does not update it anymore.

As we can note, in all the contexts, the display is active for about half of the time.
The \emph{Doze} mode has been recorded only in four contexts, \texttt{Free Time}, \texttt{CNR meeting}, \texttt{Home}, and \texttt{School}, and, on average, it represents the 4\% of the total samples collected in each context.
Finally, the \emph{Doze suspended} mode represents, on average, only the 0.4\% of the data samples, and, similarly to the previous modality, it has been recorded only in the first four contexts.

\subsection{Audio setting}

The level of the smartphone ringtone may indicate the situation in which the user is currently involved.
For example, when we are in a context in which the ringtone or notification alarm might disturb other people (e.g., during a meeting), we typically set our smartphone in silent or vibration mode.

\begin{figure}[t]
    \centering
    \includegraphics[width=\textwidth]{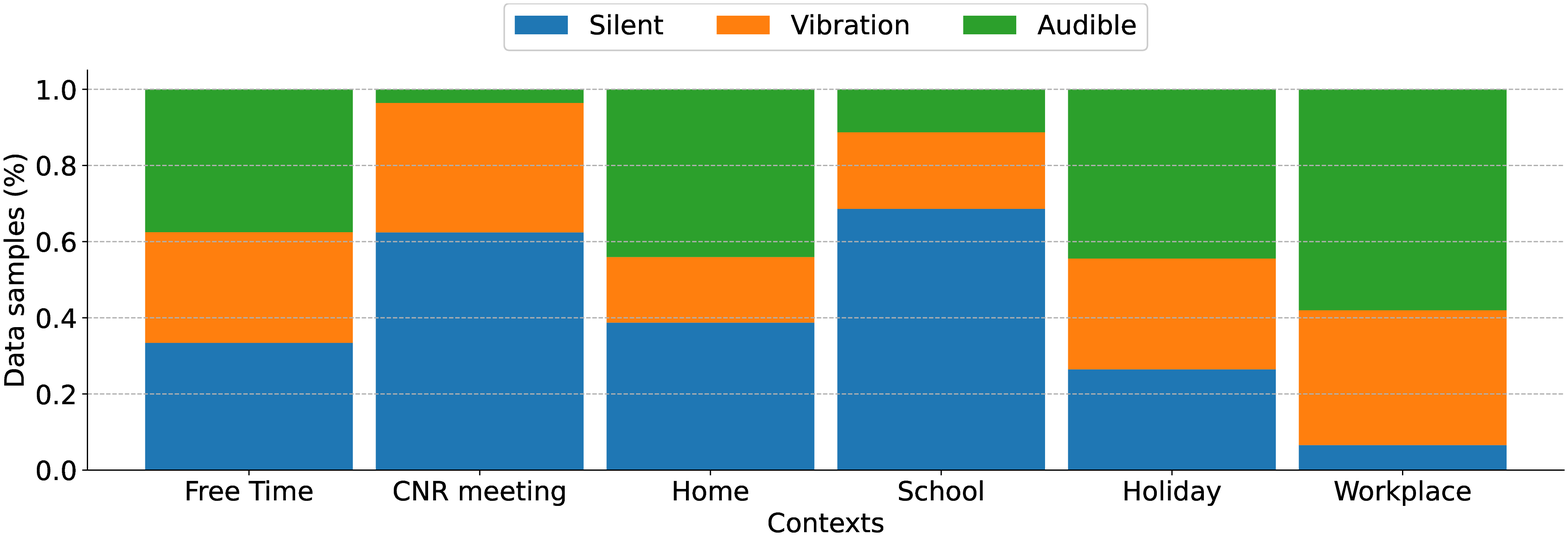}
    \caption{Audio setting in different situations.}
    \label{fig:audio_setting}
\end{figure}

Figure~\ref{fig:audio_setting} shows the amount of data samples contained in the MDF dataset that corresponds to each of the possible audio settings (i.e., \emph{Silent}, \emph{Vibration}, and \emph{Audible}), in the different contexts.
According to the audio setting, the contexts \texttt{Free Time}, \texttt{Home}, and \texttt{Holiday} are very similar to each other, where the \emph{Audible} status corresponds to about the 60\% of the total data available for those situations.
On the other hand, in a more quiet situation as in \texttt{CNR meeting} and \texttt{School}, the users' smartphones have been set in either \emph{Silent} or \emph{Vibration} modes for most of the time; while the \texttt{Workplace} scenario is characterized by a 58\% of the data that corresponds to the \emph{Audible} status.

Clearly, taking into account just the ringtone status is not enough to precisely recognize the current user's context; however, combining it with other context information can be a good approach to build an effective context-recognition system.

\subsection{Mobile applications}
\label{sec:data_analysis_apps}

Analyzing the type of mobile applications used by a person can reveal useful insights about both her preferences and the contexts she experiences during the day.
On the Google Play Store, each application belongs to a category that describes its main functionality (e.g., video game, photo editing, or shopping).
Therefore, as part of the data augmentation process of MDF, we downloaded the category information of the mobile applications used by the volunteers during the sensing experiment.

\begin{figure}[t]
    \centering
    \includegraphics[width=\textwidth]{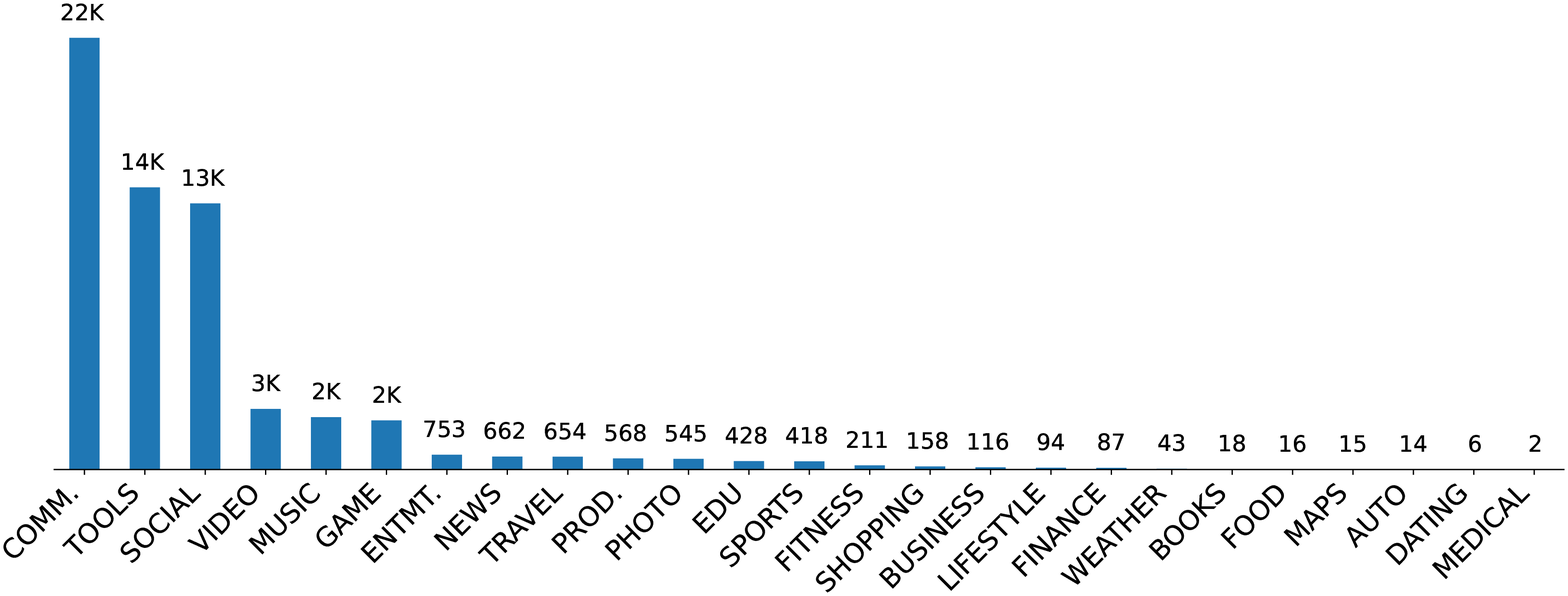}
    \caption{Number of data samples per running application category.}
    \label{fig:running_apps}
\end{figure}

Figure~\ref{fig:running_apps} shows the categories of the applications used by the volunteers during the data collection campaign.
More precisely, for each category, the figure shows the number of data samples related to the running applications categories.
The distribution presents a typical long-tail shape, where few categories have a high usage rate, while the great majority of them are used less frequently by the users.
The most used applications belong to the following categories: \emph{Communications} (\emph{COMM.}), which refers to messaging applications (e.g., WhatsApp, Telegram, and Facebook Messanger); \emph{Tools}, general tools (e.g., file explorers, dialer, and applications launcher); and \emph{Social}, and OSN client applications (e.g., Facebook, Twitter, and Instagram).
On the other hand, the least used applications are those related to different categories, including \emph{Business} (e.g., word editors), \emph{Books} (e.g., e-book readers), and \emph{Maps} (e.g., navigation).

\begin{figure}[t]
    \centering
    \includegraphics[width=\textwidth]{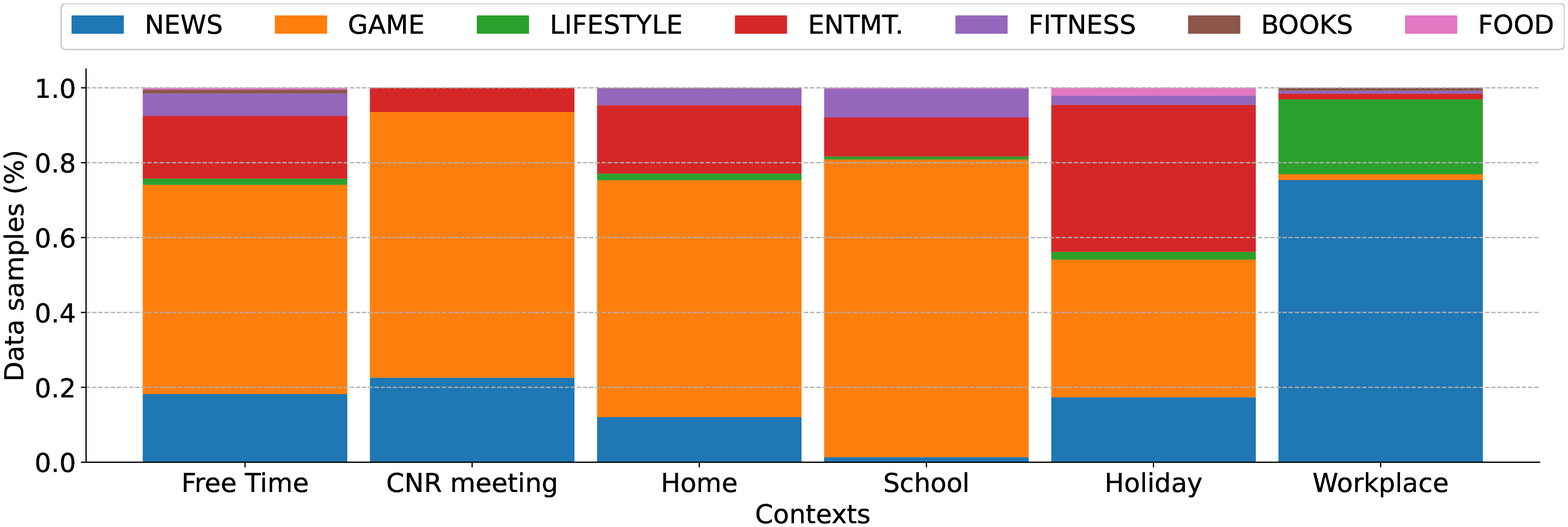}
    \caption{Amount of running application types in different contexts.}
    \label{fig:running_apps_contexts}
\end{figure}

Since the most popular applications might be used independently from the user's activity, the applications lying in the distribution tail might be more relevant to recognize the users' context.
For example, Figure~\ref{fig:running_apps_contexts} shows the percentage of data samples related to 7 of the least frequently used applications in different contexts.
As we can note, there is a clear distinction between \texttt{Workplace} and the other contexts.
The majority of the data samples generated in this situation is related to two main categories, \emph{NEWS} and \emph{LIFESTYLE}, while they represent a very small percentage (i.e., between 0\% and 23\%) in the other contexts.
Applications related to restaurants (\emph{Food} category) are only present in the \texttt{Holiday} context, while \texttt{School} is the situation with the lowest percentage of \emph{News} applications.
Finally, the contexts \texttt{Free Time} and \texttt{Home} are very similar in terms of application categories, but the latter presents a lower percentage of data samples related to \emph{FITNESS} and \emph{NEWS} applications.

\subsection{Proximity information}

Modern smartphones are equipped with different communication technologies, including Wi-Fi, Wi-Fi Direct, and Bluetooth.
Even though they have been specially designed for the creation of wireless communication channels (especially device-to-device), they also represent a useful source of data to characterize the surrounding environment.
Both Wi-Fi and Bluetooth protocols implement a specific function to identify other devices in proximity, i.e., the \emph{scan} functionality.

The analysis of proximity data can be used to model both the users' context and the physical interactions among them.
For example, the presence of specific Wi-Fi Access Points and wireless printers may indicate that the user is in the office, while the discovery of a smart TV and personal home assistants (e.g., Google Home or Amazon Alexa) represents the fact that the user is at home.
Furthermore, personal mobile devices frequently discovered during working hours may represent the smartphones of the user's colleagues, while the user's relatives can be represented by the devices that are typically in proximity during the night.

\begin{figure}[t]
    \centering
    \includegraphics[width=\textwidth]{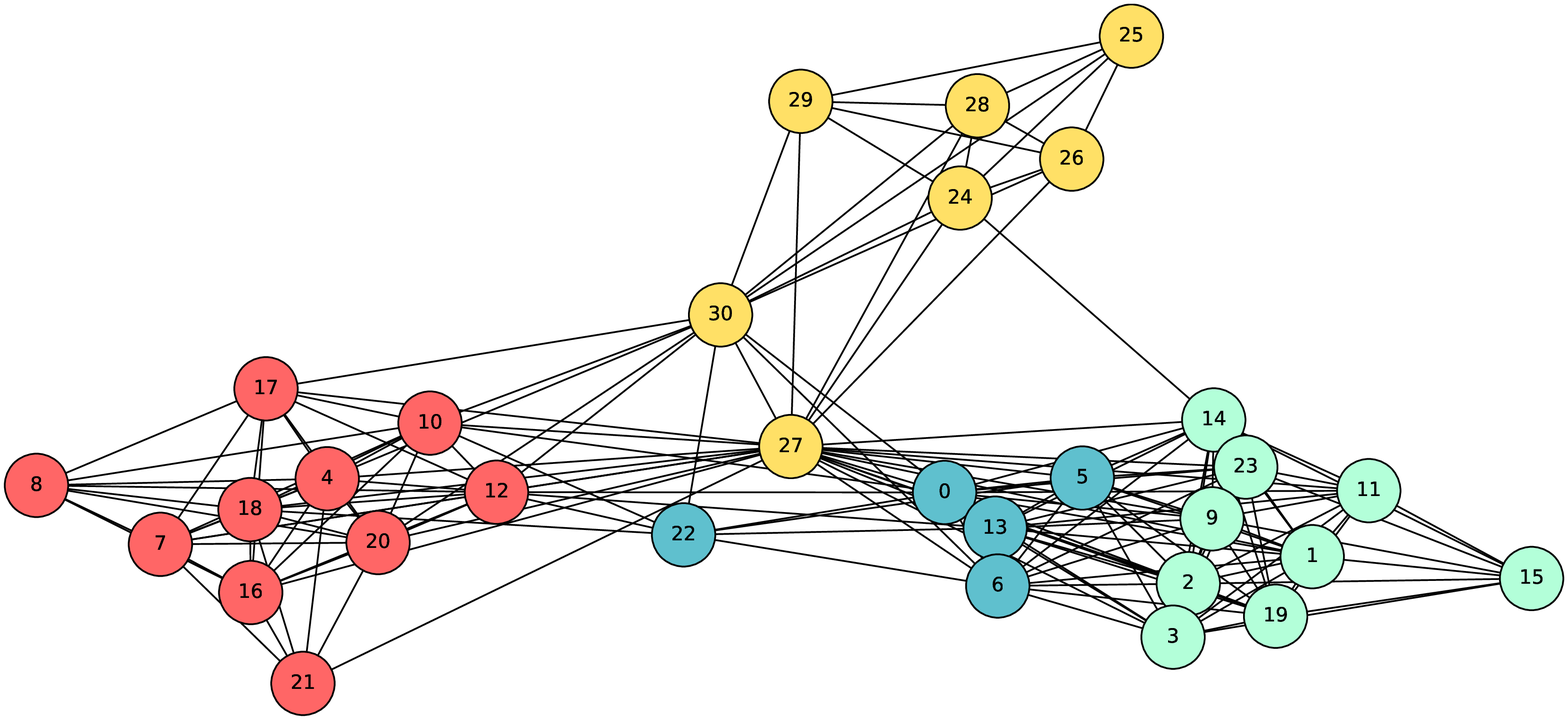}
    \caption{Proximity graph based on both Wi-Fi Direct and Bluetooth data.}
    \label{fig:proximity}
\end{figure}

Figure~\ref{fig:proximity} shows the graph built by using both the Wi-Fi Direct and Bluetooth data contained in the MDF dataset.
Nodes represent our volunteer users, while an edge connecting two nodes represents the fact that the two users were in proximity during the data collection.
Moreover, we have colored the nodes according to their category: yellow for Ph.D. students and researchers, while red, blue, and light blue represents the students from the three different high-schools involved in the sensing experiment.
The four communities are fully observable: people within the same community are more connected than users who belong to different groups.

Besides, we can note that nodes 30, 27, and 22 have connections among the four communities.
These links have been generated during the meetings organized for the start and the end of the data collection campaign, that is, when we met the students to inform them about the goals of the sensing experiment, and when we evaluated the collected data.
From a networking point of view, users 30, 27, and 22 represent examples of \emph{relay nodes} that can be exploited to disseminate data within different communities at the edge of the Internet through the use of D2D wireless communications.

\subsection{Online Social Network data}

Online Social Networks (OSNs) are virtual communities that allow people to connect and interact with each other on different topics. Users can share contents (e.g., text and multimedia files) with their friends/followers, and interact with those uploaded by the others through the use of comments and reactions (e.g., likes) to express their opinion and mood.
Therefore, OSNs represent an invaluable source of data to model both the user’s social relationships in the virtual world and her preferences.

In recent years, users have been introduced to many OSNs, where each of them focuses on a specific type of content or purpose.
For example, Twitter is mainly used to publish short opinions on different types of events (e.g., political, news, and academia), where users adopt Instagram to share photos and videos of their leisure activities.
Among the different OSNs, Facebook is one of the most used by our volunteers because it allows them to access a great variety of content.
On this platform, a user can access a timeline, where she can read the statuses (also called ``posts'') shared by her friends and interact with them. Moreover, a user can follow the status updates of different public profiles, specialized on different topics (e.g., movies or tv shows).

\input{fig_data_analysis_osn}

Figure~\ref{fig:osn_shared_items} and Figure~\ref{fig:osn_liked_items} respectively show the Empirical Cumulative Distributions (ECDF) of the type of contents shared by the volunteers on Facebook, and the public pages they follow.
As we can note, most of the users in our dataset have mainly shared generic posts and photos, while video contents are quite rare.
Moreover, they mainly prefer generic public profiles (here simply called ``Pages'') and groups related to music artists instead of other topics (e.g., books, and tv-shows).
This information can be used to both characterize the users' tastes and to infer similarities among them to optimize different downstream applications, including recommender systems, data forwarding protocols, and predicting new social interactions among people.

\section{Potential applications}
\label{sec:applications}

The MDF dataset contains a great variety of information related to people's daily activities and their social relationships in a mobile setting.
This characteristic makes MDF an invaluable source of data to create intelligent systems specially designed for the edge of the Internet.
In this scenario, the user context might rapidly change  due to mobility, daily-life activities, environmental change or discovery of new services or devices in proximity. Consequently, the ability of mobile devices to locally process sensors data is fundamental to make the system able to quickly adapt its behavior to the current context. This ability, in practice, can be represented by a single elaboration process integrated in the final user application, or by a middleware platform aimed at implementing different context processing and reasoning to support third-party applications, as detailed in the following examples. 

In this section, we present three potential context-aware applications whose reasoning process is based on  machine learning algorithms applied to the MDF dataset. Specifically, we present social link prediction and user context recognition tasks, which can represent a sort of background services to support final user applications, and a context-aware recommender systems for mobile devices, which can represent a higher layer service.

\subsection{Social links prediction}
\label{sec:social_link_prediction}

At the edge of the Internet, mobile devices can exchange data through the use of their wireless capabilities, establishing self-organizing networks based on Device-to-Device (D2D) transmission channels.
In this scenario, contents are forwarded across the network by exploiting the movements of the nodes themselves, which carry messages while waiting to enter in reciprocal radio range with either the message destination or a node that is more suitable to reach the destination.
Therefore, the ability to predict future contacts among the users can lead to an increase in the performance of data dissemination and forwarding algorithms~\cite{s19020396}.

In the social networks research field, the problem of predicting new interactions among people is typically addressed as a link prediction task (LP)~\cite{Hasan2011}.
Users' social relationships are modeled as a graph in which nodes represent users and edges characterize the existing social ties among them.
LP aims to estimate the probability to create new edges between nodes not connected in the input graph, thus, in other words, the formation of new social relationships and interactions among the users.
Specifically, we compare the performance of different LP approaches in predicting future social relationships among the users in the MDF dataset, by using information related to both physical and virtual social interactions.

\subsubsection{Data modeling}

We model the users' social relationships as an undirected and weighted social graph, where the weights of the edges represent the strength of the social ties among the users.
Formally, the social graph is defined as a triplet $G = (U, E, w)$, where $U = \{ 1,\ldots,n \}$ is the set of users, $E \subseteq U \times U$ defines the existing edges among the user nodes, and $w: E \rightarrow \mathbb{R}$ is a function that associates a weight to each edge.

Based on the data considered by the weighting function, we can define two different social graphs: the \emph{physical social graph}, and the \emph{combined social graph}.
The physical social graph models the users' face-to-face interactions, based on their proximity information.
More specifically, the weighting function of the physical social graph, $w_p(i, j)$, is defined as follows:

\begin{equation}
    w_p(i, j) = C_p(i, j),
\end{equation}

where $i, j \in U$ are two user nodes in the graph, and $C_p(i, j)$ represents the frequency of the physical contacts between them.
This is defined as the ratio between the number of times the two users have been in the proximity of each other and the total number of devices they have discovered through the use of both Bluetooth and Wi-Fi Direct scans.
Formally, the physical contacts frequency between user $i$ and user $j$ is defined as follows:

\begin{equation}
    C_p(u, i) = \frac{C(u, P_i) + C(i, P_u)}{|P_u| + |P_i|},
    \label{eq:contact_frequency}
\end{equation}

where $P_i$ is the list of physical contacts of $i$, $C(u, P_i) = \sum_{s \in P_i} \mathbbm{1}_{\{ s = u\}}$ is the number of occurrences of $u$ in $P_i$, and $|P_i|$ is the total number of physical contacts of $u$.

On the other hand, the combined social graph models the users' relationships by taking into account a heterogeneous set of information, including not only their physical contacts but also their affinity in the virtual world.
Formally, we define the weighting function of the combined social graph, $w_c$, as follows:

\begin{equation}
    w_c(u, i) = \alpha \cdot w_p(u, i) + (1-\alpha) \cdot aff(u, i),
    \label{eq:combined_weight}
\end{equation}

where $aff(u, i)$ represents the affinity between the two users, and $\alpha \in [0,1]$ is a tunable parameter to weight the importance of the physical and virtual data.
We define the users' virtual affinity as the mean value of the following features: (i) virtual interactions frequency on the OSN (i.e., comments and reactions on shared posts, videos, and photos), phone calls frequency, their friendship status on Facebook (i.e., whether they are friends or not), and their similarity in preferences.
Both the virtual interactions and phone calls frequencies are calculated using Equation~\ref{eq:contact_frequency} but with different information (i.e., OSN activities and phone calls), while the similarity in preferences between two users is defined as the Jaccard Index between the sets of their most-used mobile applications.
The rationale behind this is that mobile application usage can partially characterize the user's preferences.
According to the homophily principle, people with similar interests tend to create social relationships among them~\cite{doi:10.1146/annurev.soc.27.1.415}.
Therefore, using people preferences as an additional feature of our model can lead to more precise predictions in terms of the creation of new social links among the users.

\subsubsection{Experiments}

To evaluate the effectiveness of the proposed social model we follow the standard LP pipeline~\cite{alex2019evalne}.
First, we randomly split the social graph edges into two distinct sets, $E_{train}$ and $E_{test}$, ensuring that the training edges represent a single connected component.
We fix the size of $E_{train}$ and $E_{test}$ as the 80\% and 20\% of the original graph size, respectively.
Since existing edges represent only positive examples, we also randomly generate sets of both train and test non-edges, $D_{train}$ and $D_{test}$, respectively.
Any nodes pair $\{i, j\} \notin E_{train}$ is considered as a valid train non-edge example, while samples in $D_{test}$ are randomly selected from $E \cup D_{train}$.
Moreover, in order to avoid creating imbalanced train and test sets (i.e., with different numbers of positive and negative examples), we use the same number of existing and non-existing edges for both sets.

Then, we use the training dataset $E_{train} \cup D_{train}$ to train LP methods and finally evaluate their predictions by using the test set $E_{test} \cup D_{test}$ as Ground Truth.
Among the several approaches proposed in the literature for the LP task, we have selected the following 8 state-of-the-art solutions.

\noindent \emph{Baseline heuristics.}
Traditional LP heuristics use the neighborhood $\Gamma(i)$ and $\Gamma(j)$ for each node pair $\{i, j\}$ to calculate a similarity score that can be interpreted as the creation probability of a new link among them.
We consider the following heuristics: \emph{Common Neighbors} (\emph{CN}) defined as $CN(i,j) = |\Gamma(i) \cap \Gamma(j)|$; \emph{Resource Allocation Index} (\emph{RAI}), $RAI(i,j) = \sum_{k \in \Gamma(i) \cap \Gamma(j)} 1/\Gamma(k)$; \emph{Jaccard Index} (\emph{JI}), $JI(u, i) = |\Gamma(i) \cap \Gamma(j)|/|\Gamma(i) \cup \Gamma(j)|$; \emph{Adamic Adar Index} (\emph{AAI}), $AAI(u,i) = \sum_{k \in \Gamma(i) \cap \Gamma(j)} log|1/\Gamma(k)|$; and \emph{Preferential Attachment} (\emph{PA}) which is defined as $PA(i, j) = |\Gamma(i)| \cdot |\Gamma(j)|$.
In addition, we also test a new method proposed by Mara et al.~\cite{alex2020network} called \emph{Node Embedding Heuristic} (\emph{NEH}) that joins the aforementioned heuristics in a 5-dimensional feature vector, followed by logistic regression to obtain link predictions.
    
\noindent \emph{Node Embeddings.}
Node Embeddings (NE) refer to a set of techniques that learn low-dimensional representations of the network nodes, called embeddings. 
The embeddings are feature vectors that can be used as input for downstream machine learning applications, including the LP task~\cite{8395024}.
Among the different proposals, we have selected two of the most popular NE solutions based on random walks, \emph{DeepWalk} (\emph{DW})~\cite{10.1145/2623330.2623732} and \emph{Node2Vec} (\emph{N2V})~\cite{10.1145/2939672.2939754}.
These algorithms use random walks to determine similarities among the nodes.
Then, they use the generated random walks as the input of the Skip-Gram model~\cite{mikolov2013efficient} to generate the final node embeddings.
The main difference between DW and N2V relies on the approach used to generate the random walks.
While the former does not take into account the edge weights and uses random walks with fixed transition probabilities, the latter biases the random walks according to edge weights and control the direction of the walkers by using two parameters: the return parameter $p$ to go back towards the previous node, and the in-out parameter $q$ to perform a forward step.
NE produces a low-dimensional representation of the graph nodes.
However, in order to perform the LP task, we need to represent the edges.
According to the literature, we tackle this problem by using the Hadamard operation $(x_i \cdot x_j)$, thus representing the edge between nodes $i$ and $j$ as the dot product between their corresponding embeddings, $x_i$, and $x_j$.
Finally, we use the result as an input of logistic regression, and we compare the predicted links with the Ground Truth to evaluate the performance of the NE methods.

Besides the heuristics and NE methods, we propose an alternative model for LP, based on deep neural networks (\emph{NN}).
Specifically, we cast the LP task into a binary classification problem, where existing edges are associated with the label $1$, while the label $0$ represents non-edges in the original graph.
Therefore, NN learns the non-linear function $f(x;\Theta) = \hat{y}$, where $\hat{y}$ represents the predicted label, $\Theta$ is the set parameters learned by using the Adamax algorithm~\cite{kingma2014adam}, and $x$ is the input features vector.
For each pair of nodes $\{i,j\}$, we define the corresponding input vector $x$ as the combination of the 5 features used in Equation~\ref{eq:combined_weight} to calculate the edge weights in the combined social graph: (i) the users' physical contacts frequency, (ii) their virtual interactions frequency, (iii) phone calls frequency, (iv) their friendship status, and (v) their similarity in preferences.

To define the ANN architecture we follow the traditional approach for binary classification.
We use the \emph{Rectified Linear Unit} (\emph{ReLU}) as activation function for the hidden layers, and the output layer uses the \emph{Sigmoid} activation function to calculate the predicted label $\hat{y}$.
To evaluate the predicted label and then update the model parameters we use the \emph{Binary Cross-Entropy} loss function that calculates the discrepancy between the actual and predicted labels~\cite{janocha2017loss}.
Moreover, to speed-up the learning process, we use \emph{Batch normalization}~\cite{ioffe2015batch} that normalizes and scales the activation output of the hidden layers, and we rely on the \emph{Dropout}~\cite{NIPS2013_4878} technique to randomly deactivate hidden units during the training, thus avoiding the overfitting of the final model.

\subsubsection{Results and discussion}

To fairly compare the performance of the considered algorithms we measure the LP accuracy in terms of \emph{Area Under the Receiver Operating Characteristics} (\emph{AUROC}).
AUROC describes the binary classification performance of a model comparing the true positive rate with the false positive rate, at various threshold settings~\cite{FAWCETT2006861}.
The AUROC is defined in $[0,1]$, where $0.5$ corresponds to the accuracy of a random guesser (i.e., a model that provides random classification) and $1$ represents the perfect match between the predictions and the Ground Truth.

\input{social_lp_parameters_tuning}

The performances of both NE and ANN depend on the values of several hyperparameters.
For this reason, we first find the best configuration for each model, by adopting the Grid search strategy.
Table~\ref{tab:social_link_prediction_tuning} shows the search space for each parameter and highlights in boldface the values that led us to the best results.
For N2V we have specified the best hyperparameters that we have found for both graphs, physical (\emph{p}) and combined (\emph{c}) social graph.
This is due to two main reasons.
Firstly, DW, as well as the simple heuristics, takes into account only the topology of the graph, ignoring the possible weights associated with the edges.
In other words, for these methods, there is no difference between the physical and social graphs.
Secondly, ANN is only applicable to the combined social graph, due to the input features we have previously defined for this model.

DW and N2V share some common hyperparameters: \emph{dimensions} refers to the number of dimensions of the output embeddings, \emph{number-walk} and \emph{walk-length} are respectively the number and length of the random walks performed by the algorithms to define the similarity between two nodes, and \emph{window-size} is Skip-Gram parameter to build the embeddings starting from the input random walks.
Moreover, the N2V parameters \emph{p} and \emph{q} are used to bias the random walkers direction.

The ANN hyperparameters that we have evaluated are the following: \emph{layers}, the number of hidden layers; \emph{hidden-units}, the number of neurons in each hidden layer; \emph{learning-rate}, the learning rate value to use with the Adamax optimization algorithm; \emph{dropout-rate}, that represents the percentage of neurons to randomly deactivate during the model training; \emph{batch-size}, the number of samples that will be propagated through the network at each training step; and \emph{epochs}, the number of training iterations.

Table~\ref{tab:social_link_prediction_results} shows the LP accuracy in terms of AUROC scores obtained by the reference algorithms.
LP heuristics obtain similar results (i.e., about 0.88), excluding PA and NEH whose scores are 0.67 and 0.83, respectively.
NE embedding methods underperform the simpler heuristics, confirming similar results obtained in the literature~\cite{alex2020network}.
N2V gets better performance compared with DW (i.e., 0.84 instead of 0.81), showing a slight improvement with the combined social data.
The Deep Learning model ANN overcomes the other approaches, obtaining a LP accuracy of 0.95.
This result is surely due to the ability of the model to learn effective non-linear combination of the input data, but it also proves the great advantage of modeling the users' social relationships by integrating both physical and virtual social interaction data.
This can be achieved only by combining physical interactions with Online Social Network data, demonstrating on the one hand the need of collecting this type of information from personal mobile devices, and on the other hand the need for the MDF dataset, which is unique in terms of such heterogeneous data collected by real users in the wild.
\input{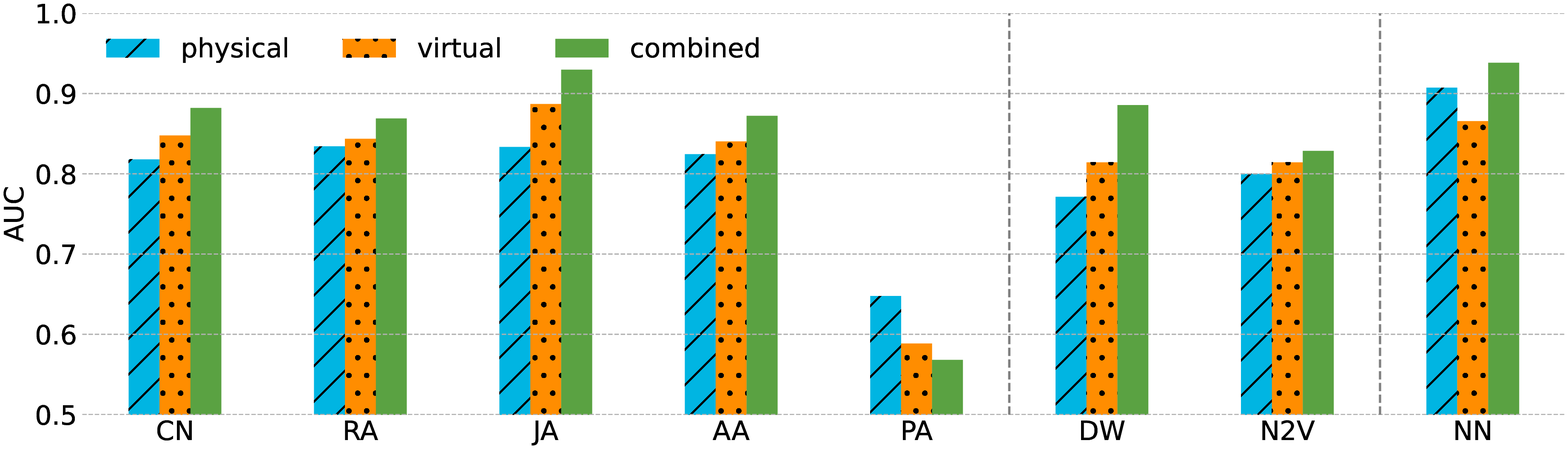}

\subsection{Daily-life context recognition}

The automatic recognition of the user context, based on a great variety of heterogeneous sensor data, is fundamental to optimize and personalize pervasive computing applications at the edge of the Internet.
Network optimization~\cite{7462489}, device adaptation~\cite{MIZOUNI20147549}, and augmented reality~\cite{10.1145/3393672.3398640}, are only some of the possible application scenarios.
In addition, since collected data include sensitive information regarding the user behaviour, habits, preferences and personal data, the execution of this process directly at the edge ensures privacy and security methodologies, which contribute to increase both the system's trustworthiness and the user awareness in the data ownership.

User context recognition inherits important contributions from the activity recognition research field (mainly focused on  movements and body posture~\cite{MORALES2017388}) and further extends the research activity by modeling more abstract concepts (e.g., daily-life situations) combining together information generated by heterogeneous data sources~\cite{6985718}.

Like the previous example, this task can represent a sort of background service aimed at supporting upper-layer applications that can benefit from this feature. This service must be able to learn how to identify different user contexts in a data-driven way, and to this aim it leverages on a combination of heterogeneous information extracted from the MDF dataset.

\subsubsection{Data modeling}
\label{sec:context-rec_data}

Our context recognition system is based on the combination of 7 different sensor information available on commercial smartphones: \emph{user gait}, \emph{audio state}, \emph{charging state}, \emph{display status}, \emph{user location}, \emph{weather conditions}, and \emph{Wi-Fi status}.
For a given minute, the system samples measurements from these seven sensors, and the final goal is to predict the most appropriate label that better describes the daily-life situation in which the user is currently involved.

\begin{figure}[t]
    \centering
    \includegraphics[width=\textwidth]{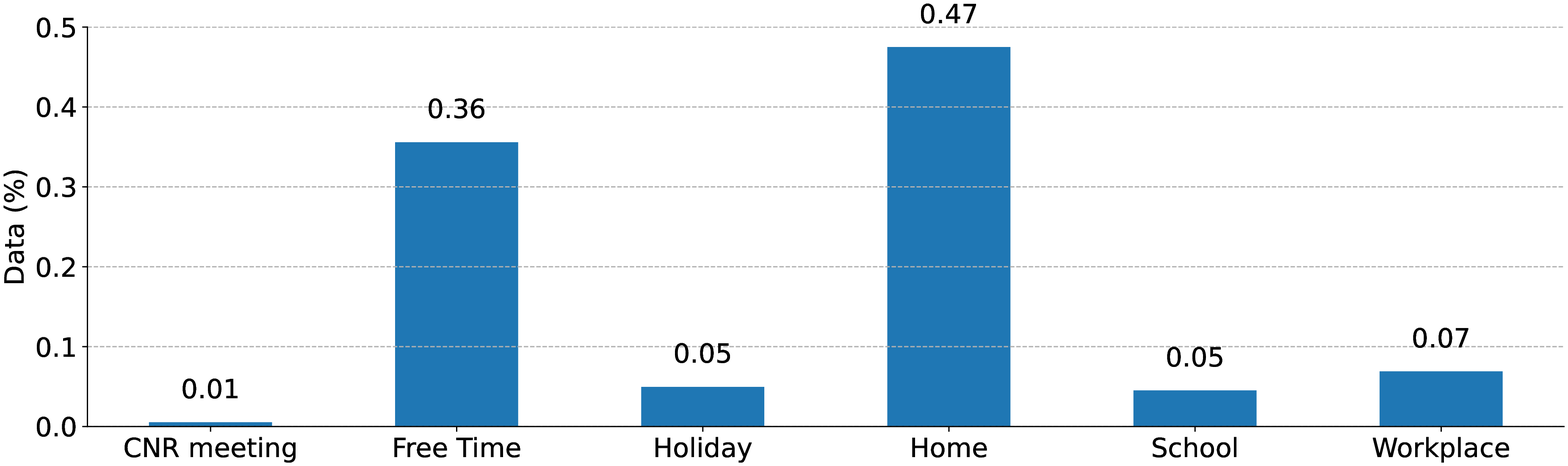}
    \caption{Labels distribution of the context recognition dataset.}
    \label{fig:context_recognition_dataset}
\end{figure}

In order to train the context recognition system, we have created a dataset of 67575 data samples, where each data sample is a high-dimensional vector composed of a total of 100 features extracted from the considered sensors and described in Section~\ref{sec:data_analysis}.
Each data sample represents 1-minute of the context information contained in the MDF dataset, and it is associated with the closest context label that we have defined in Section~\ref{sec:data_augmentation}: \texttt{Home}, \texttt{School}, \texttt{Workplace}, \texttt{CNR meeting}, \texttt{Free Time}, and \texttt{Holiday}.
Figure~\ref{fig:context_recognition_dataset} shows the percentage of data samples associated with each label.
As we can note, the dataset is highly imbalanced, where the 83\% of the data samples is labeled with \texttt{Home} and \texttt{Free Time}, and the remaining 17\% is distributed among the other 4 classes.

\subsubsection{Experiments}

We compare the performance of two different types of classifiers: Decision Tree (DT) and a Deep Neural Network model (DNN).
DT is a tree-based prediction model, where leaves represent the target classes, and branches represent conjunctions of input features that lead to those class labels.
The main goal of DT is to automatically learn the branch splits that best fit the input data.

Similar to what we have done for the social link prediction problem, we have implemented the DNN model with multiple hidden layers, using RELU as activation function to let the neural network to learn a non-linear mapping between the input features and the output classes.
Also in this case, we have used Batch normalization to speed-up the learning process and the Dropout technique to avoid the model overfitting.
The main difference between the two neural networks consists in the type of classification problem: while in the social link prediction the neural network must predict one of the two possible labels (i.e., link or non-link), in this case, our model must be able to handle a multi-label classification problem.
For this reason, we have implemented the output layer by using the \emph{Softmax} activation function with 6 output units, one for each of our considered context labels, and we have used the \emph{Categorical Crossentropy} loss function to update calculate the classification error and, then, update the model weights by using the Adamax, accordingly.

\subsubsection{Results and discussion}

\input{context_recognition_parameters}

As a first step, we have split the entire dataset into training (80\%) and test (20\%) sets.
We have used the former set to both train and optimize the best models, while the latter has been used to evaluate the performance of the two approaches.
Table~\ref{tab:context_recognition_parameters} shows the hyperparameters tuned by using the Grid-search approach and highlights in bold the best values.
The most important parameters of DT are the following: \emph{max-depth}, that represents the maximum depth of the decision tree; \emph{min-sample-leaf}, the minimum number of training samples used to define a leaf node; and \emph{min-samples-split}, the minimum number of training samples required to create a new branch.

Compared to the previous DNN, in this case, we need a more complex architecture to implement the multi-label classification function.
In fact, according to Grid-search results, we have obtained the best performance with 3 hidden layers, each of them composed of 200 hidden units.
Moreover, to obtain a model that generalizes well with data samples never seen before, we need to randomly deactivate the 20\% of the neurons during the training phase (i.e., by setting 0.2 as \emph{dropout-rate}).

\input{fig_context_recognition_results}

As we noted in Section~\ref{sec:context-rec_data}, the distribution of the labels contained in MDF is highly skewed, where some classes appear more frequently than others. This characteristic is particularly common in real-world datasets, but it can negatively influence the generalization and reliability of machine learning algorithms, since the resulting classifiers may be biased by the majority classes~\cite{Krawczyk-learning}.
Therefore, to deal with this issue, we compare the performance of the considered solutions in terms of AUROC, which is an effective metric to evaluate classifiers with imbalanced datasets~\cite{BRADLEY19971145, 7835710}.
In terms of AUROC, the best models obtained the following results: \textbf{0.945} for DT, and \textbf{0.999} for DNN.
The deep model clearly overcomes DT, providing a nearly perfect classification.
This result is also proved by the confusion matrices generated by using the predictions of the two classifiers.
Figure~\ref{fig:context_recognition_results} shows the graphical representations of the two confusion matrices, where we have encoded the context labels as follows: $0$ represents the class \texttt{CNR meeting}, $1$ is \texttt{Free Time}, $2$ is \texttt{Holiday}, $3$ is \texttt{Home}, $4$ is \texttt{School}, and $5$ denotes the \texttt{Workplace} context.
As we can note from Figure~\ref{fig:cr_DT_results}, DT fails in recognizing the contexts \texttt{CNR meeting} and \texttt{Holiday}, while it scores well for the other classes.
The main confusion for DT is related to similar contexts.
In fact, \texttt{CNR meeting} and \texttt{Workplace} refer to situations that occurred in the same location (i.e., the CNR research center), while the main difference between \texttt{Holiday} and \texttt{Free Time} relies on the geographical location, that is, \texttt{Holiday} refers to situations occurred outside the geographical bounding box shown in Figure~\ref{fig:map_locations}.

On the other hand, the classification results show in Figure~\ref{fig:cr_NN_results} prove that DNN can correctly recognize all the considered context situations.
Similarly to DT, DNN shows little confusion with the \texttt{Holiday} label, classifying 9\% of the data samples as \texttt{Free Time}.
To further improve the classification performance, one should also include additional features such as the distance between the user's home and her current location.
This, as the inclusion of additional sensors, would be one direction for future work.

\subsection{Pervasive context-aware recommendations}

At the edge of the Internet, data transmission leverages both the Internet core and direct communications among devices in proximity.
In this scenario, users can potentially access a massive amount of data coming from both remote or edge servers, and other devices in proximity.
Moreover, the actual utility of contents available at the edge typically depends on the current situation in which the user is involved~\cite{CONTI201851}.
Therefore, personal mobile devices must be able to assist their users in discovering and selecting the most relevant content, according to the users' context and needs.

One of the most promising approaches is represented by the use of a new emerging type of Context-Aware Recommender Systems (CARS), called \emph{Pervasive CARS} (\emph{P-CARS})~\cite{ARNABOLDI20173, 10.1145/3298689.3347067, mettouris2014ubiquitous}.
CARS exploit both the user's preferences (inferred from the past user-system interactions) and various information related to the user's context to provide more accurate and personalized recommendations.
However, traditional approaches rely on centralized architectures, in which the mobile device is considered as a simple source of context data (acting as a client), and both the recommendation engine and the contents are stored on the remote servers.
Actually, the mobile environment is characterized by highly dynamic contexts and not all the contents are necessarily stored on remote servers but they can be directly maintained on the local devices. In this case, considering also the mobility of both users and devices, each mobile node has a limited time to discover interesting contents and recover them. For this reason, P-CARS are directly executed on the local devices and exploit both their computational and sensing capabilities to analyze and select the most relevant contents available nearby, thus providing faster and more personalized recommendations to the users.

One of the main challenges in P-CARS research is represented by the scarce availability of public datasets containing heterogeneous and realistic data.
Most of the available datasets typically used in the CARS evaluation have been collected from traditional (i.e., centralized) multimedia and e-commerce domains, where the amount and variety of contextual information are limited, and mobile devices are considered as simple clients that access a remote catalog of contents or products~\cite{RAZA201984, Haruna2017}. In this regard, the MDF dataset represents a unique source of real-world contextual data that fully characterize the different aspects of the user's physical and social context in the mobile environment. In this section, we show how the MDF dataset can be used to successfully evaluate a new P-CARS especially designed to assist the mobile user by automatically identifying and selecting relevant contents at the edge of the Internet.

\subsubsection{Data preparation}

As a proof of concept application to highlight the advantages of context-aware recommendations at the edge, we present a simplified task aimed at recommending mobile applications based on the local user's context.
To this aim, we built a dataset $D$ where each data sample is represented by the following triplet: $\langle U, A, C\rangle$, where $U$ represents the user ID, $A$ is the application ID, and $C$ is the context in which the user was involved when she was using the mobile application.
As context information, we have used the context labels defined in Section~\ref{sec:data_analysis_annotation} and 3 additional features: (i) the application category (e.g., Communication, Social, or Video), (ii) the time of the day in which the application has been used (e.g., Morning or Afternoon), and the day of the week (e.g., Weekday or Weekend).

The information contained in the MDF dataset represents only positive examples, that is, the applications used by the volunteers during the day.
However, to effectively train a P-CARS model, we also need negative examples.
To this aim, we used the same technique employed in Section~\ref{sec:social_link_prediction}, negative sampling: we randomly created new data samples that are not already contained in the set of positive examples $D$. The resulting dataset is composed of 126034 data samples, with a perfect balance (50\%-50\%) between positive and negative examples.

\subsubsection{Experiments}

Since our dataset is composed of positive and negative samples, we can treat the recommendation problem as a binary classification task.
Similar to what we have done in the previous section, we compare the performance of two classification methods: Decision Tree (DT), and a Neural Network model (NN).
As far as NN is concerned, we used RELU activation function to learn a non-linear transformation of the input, Batch normalization technique to speed-up the learning process, and  Dropout approach to avoid model over-fitting.
In addition, since we are dealing with a binary classification problem, we used the Binary Cross-Entropy loss function to calculate the classification error and then update the model weights with the Adamax learning algorithm.

Besides the comparison of the two models, we also evaluate the importance of taking into account the context information in the recommendation task.
Specifically, for each classification approach, we create two models, called \emph{BASE} and \emph{CARS}, that use respectively the basic user-app interactions, and the context information to provide the recommendations to the user.

\subsubsection{Results and discussion}

\input{cars_parameters}

As a first step, we split the entire dataset into training (80\%) and test (20\%) sets.
The former set is used to train and optimize the machine learning models, while the latter is used to evaluate their performance against data samples never seen before.

Table~\ref{tab:cars_parameters} shows the parameters tuned with the Grid search approach, and highlights in bold the values of the best models.
In this case, the best DT model is characterized by a \emph{max-depth} of 11 branches, at least one sample to define a leaf node, and two data samples are enough to create a new branch in the decision tree.
On the other hand, a simple shallow NN composed of one hidden layer with 200 units is sufficient to fit the training set; while the random deactivation of 80\% of the hidden units allows maintaining a good generalization over unseen samples.

\input{cars_results}

Table~\ref{tab:cars_results} shows the results obtained by the best models over the test set, in terms of AUROC.
According to the results, DT has better performance than NN when the classification task is based only on the user-app past interactions, obtaining an AUROC value of 0.963, which corresponds to a gain of 2\% with respect to the accuracy level reached by NN.
On the other hand, when the context data are used as additional information, the NN performs best, approximately providing perfect recommendations, obtaining 0.994 in terms of AUROC.

\input{fig_cars_nn}

Besides the results obtained by the two models, it is worth noting that the classification accuracy substantially increases when the context information is taken into account in the recommendation process.
This becomes clearer by observing the performance of the NN during the training phase, where we have used 20\% of the training data as a validation set to tune the hyperparameters.
Figure~\ref{fig:cars_nn_loss} shows the value of the loss function during the training, while Figure~\ref{fig:cars_nn_auc} shows how the classification accuracy increase according to the number of training epochs.
As we can note, the CARS model overcomes the simpler model both in terms of loss and AUROC values, obtaining a 20\% loss gain, which corresponds to a 6\% more accurate prediction than those provided by the BASE model.

The obtained results confirm our assumption related to the relevance of the context information in mobile environments, proving that a context-aware recommender system is able to provide more accurate and personalized recommendations to the users than a standard system.
Moreover, this example shows how MDF contributes to the validation and evaluation of the proposed solution, and the effective need of public datasets to support this research area.

\section{Conclusion}
\label{sec:conclusion}

In this paper we presented MyDigitalFootprint (MDF), a large-scale dataset collected from commercial smartphones that represents a rich source of data to characterize the user physical and social context at the edge of the Internet.
Differently from datasets derived from experiments conducted in controlled environments, we performed a data collection campaign in-the-wild, guaranteeing a detailed representation of the real-world as seen by personal mobile devices.
In fact, the dataset contains a wide range of heterogeneous sensors data collected from commercial mobile devices of 31 volunteer users within a period of 2 months.
The monitored sensors include physical hardware embedded in the smartphones (e.g., GPS, gyroscope, and accelerometer), and several data sources describing the user interactions with the mobile device, her habits, and the surrounding environment.
MDF significantly differs from the other available datasets since it collects data able to describe different context's domains, i.e., the physical context and the social context of a user in terms of physical social relationships (e.g., face-to-face meeting derived from proximity information), and online social interactions derived from Online Social Network platforms.

After a detailed analysis of the collected data, we used MDF to implement and evaluate, as a proof-of-concept, three context-aware mobile applications based on machine learning techniques: (i) the prediction of physical interactions among mobile users, (ii) the recognition of daily-life context, and (iii) a pervasive context-aware recommender system.
Each application highlights different peculiarities of the data contained in the proposed dataset: while the prediction of physical contacts is based on proximity and Online Social Network data, the context recognition exploits sensing data to predict 6 daily-life activities.
Finally, the recommender system presented as proof-of-concept leverages on sensors data to suggest the most appropriate mobile application to the user, based on the situation in which she is currently involved.
The evaluation of such applications clearly shows the benefit of combining the heterogeneous context data collected in MDF to improve the effectiveness of the system, providing more accurate and personalized services to mobile users compared to baseline solutions that leverage on the limited information typically contained in public datasets.
 
The methodology we used for the data collection campaign, in combination with the large number of rich context and social data contained in our dataset, make MDF a valuable foundation for the definition and assessment of novel proposals in various research fields.
Besides the applications proposed in this paper, MDF can be used, for example, to: (i) study how activities in the cyber world affect real social interactions and vice versa; (ii) benchmark protocols and algorithms of data dissemination in self-forming wireless and mobile networks; and (iii) design mobile health (m-health) applications, based on the user's behavior and social interactions inferred from smartphone-embedded sensors.
We encourage the research community to utilize the presented dataset for their research and development goals.

Our future work will follow two main research directions.
Firstly, we plan to extend MDF by collecting a broader set of users' defined labels that characterize their daily-life activities and situations. However, manually label the smartphone sensors data is a tedious and time-consuming task, which could affect the success of the data collection campaign.
A possible solution to this issue could be the design of a back-end infrastructure that collects the smartphones data and assists the users in labeling them.
In this case, the server should execute a simple but effective unsupervised classifier (e.g., K-Nearest Neighbors~\cite{1053964}) to guess the correct label of new incoming data samples, based on their similarity with observations previously classified by other participants.
Then, the server will send the predicted labels to the originated devices, where the users can either accept or modify them, also specifying new context labels not already included in the remote classifier.
In this way, the server would lighten the users' workload, simplifying the collection of semantically rich context data from the mobile devices.

Regarding the second line of research, we are developing a  middleware solution that integrates the context-aware functionalities of the proposed proof-of-concepts in order to support heterogeneous application domains. The middleware locally processes the context data and implements the reasoning tasks to make the system able to automatically adapt to the user current situation. Even in this case, MDF represents a valuable resource to assess the performance of the middleware and the reasoning tasks. 

\bibliography{paper}

\end{document}

%% file: fig_mdf_app.tex
\begin{figure}[t]
    \centering
    \begin{subfigure}{.49\textwidth}
        \centering
        \includegraphics[width=.9\linewidth]{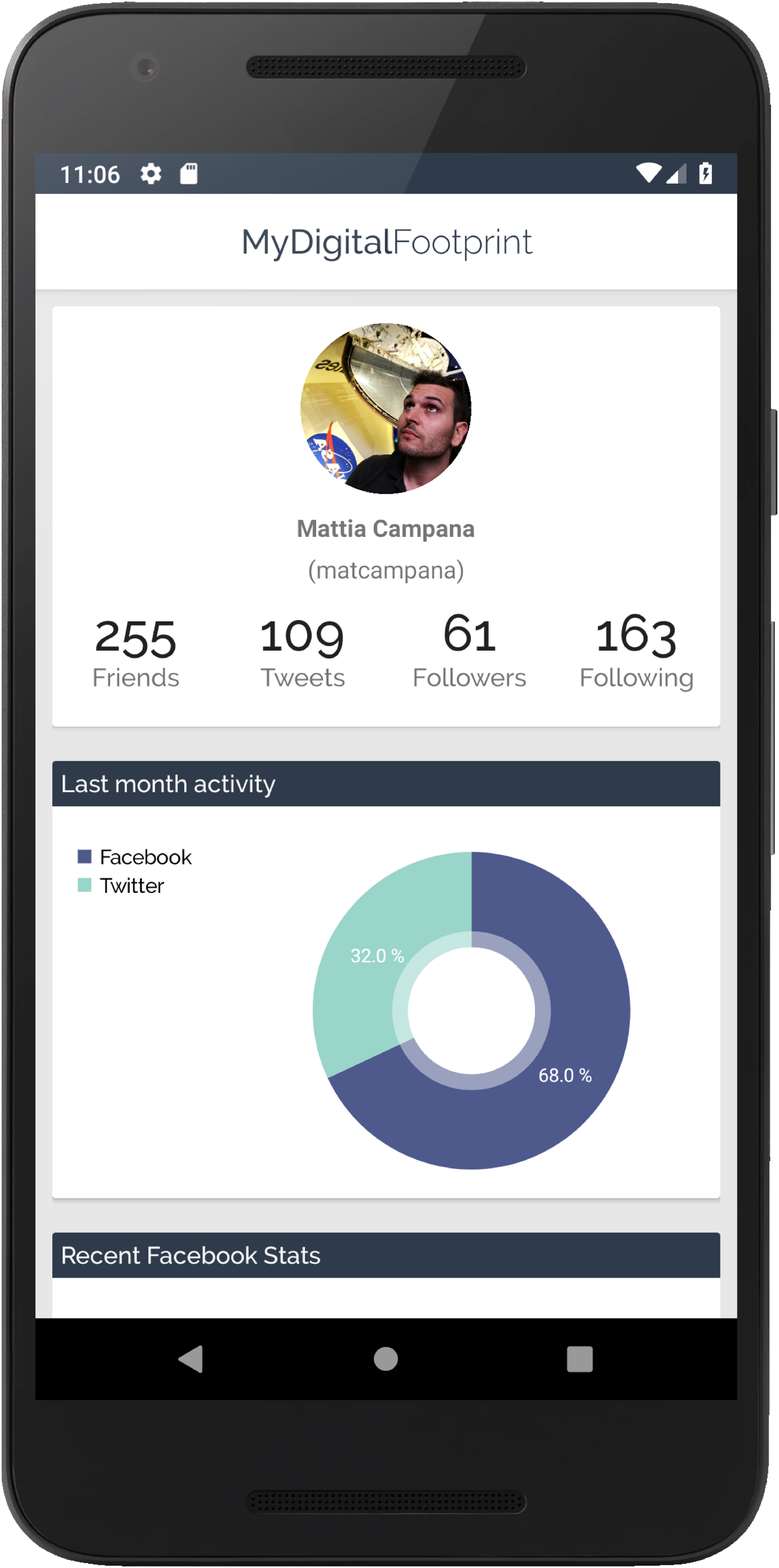}
        \caption{}
        \label{fig:android_app}
    \end{subfigure}
    \begin{subtable}{.49\textwidth}
        \footnotesize
        \begin{tabular}{p{0.5\linewidth}@{\hskip 10pt}p{0.4\linewidth}}
            \toprule
            \textbf{Sensor}  & \makecell[r]{\textbf{Sampling Rate}}        \\
            \midrule
            \makecell[l]{
                User's gait\\
                Calendar\\
                Audio\\
                Battery\\
                Display\\
                Physical sensors\\
                Mobile cells\\
                Bluetooth scans\\                          
            }                                       & \makecell[r]{1 min}   \\\midrule
            Wi-Fi P2P                               & \makecell[r]{2 min}   \\\midrule
            Wi-Fi                                   & \makecell[r]{4 min}   \\\midrule
            \makecell[l]{
                Location\\
                Running apps\\
            }                                       & \makecell[r]{5 min}   \\\midrule
            \makecell[l]{
                Apps usage\\
                Installed applications\\
                Weather conditions
            }                                       & \makecell[r]{1 h}     \\\midrule
            \makecell[l]{
                Multimedia\\
                Phone calls\\
                Bluetooth connections
            }                                       & \makecell[r]{On event}   \\
            \bottomrule
        \end{tabular}
        \caption{}
        \label{table:sensors}
    \end{subtable}
    \caption{The UI of the sensing mobile application (a), and the sampling rate used for each sensor category considered during the data collection.}
\end{figure}

%% file: fig_map.tex
\begin{figure}[t]
    \centering
    \begin{subfigure}{.49\textwidth}
        \centering
        \includegraphics[width=\linewidth]{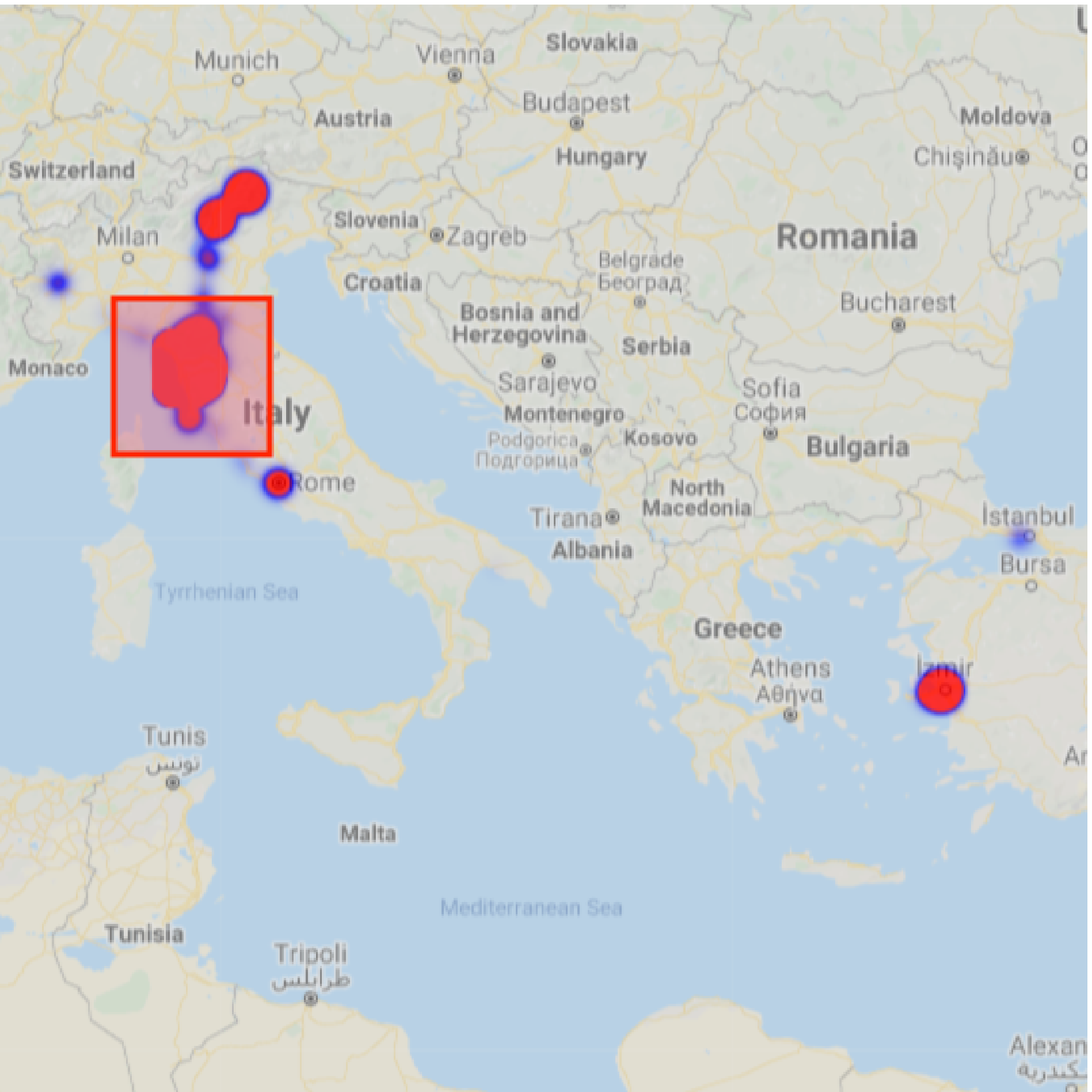}
        \caption{}
        \label{fig:loc_all}
    \end{subfigure}
    \begin{subfigure}{.49\textwidth}
        \centering
        \includegraphics[width=\linewidth]{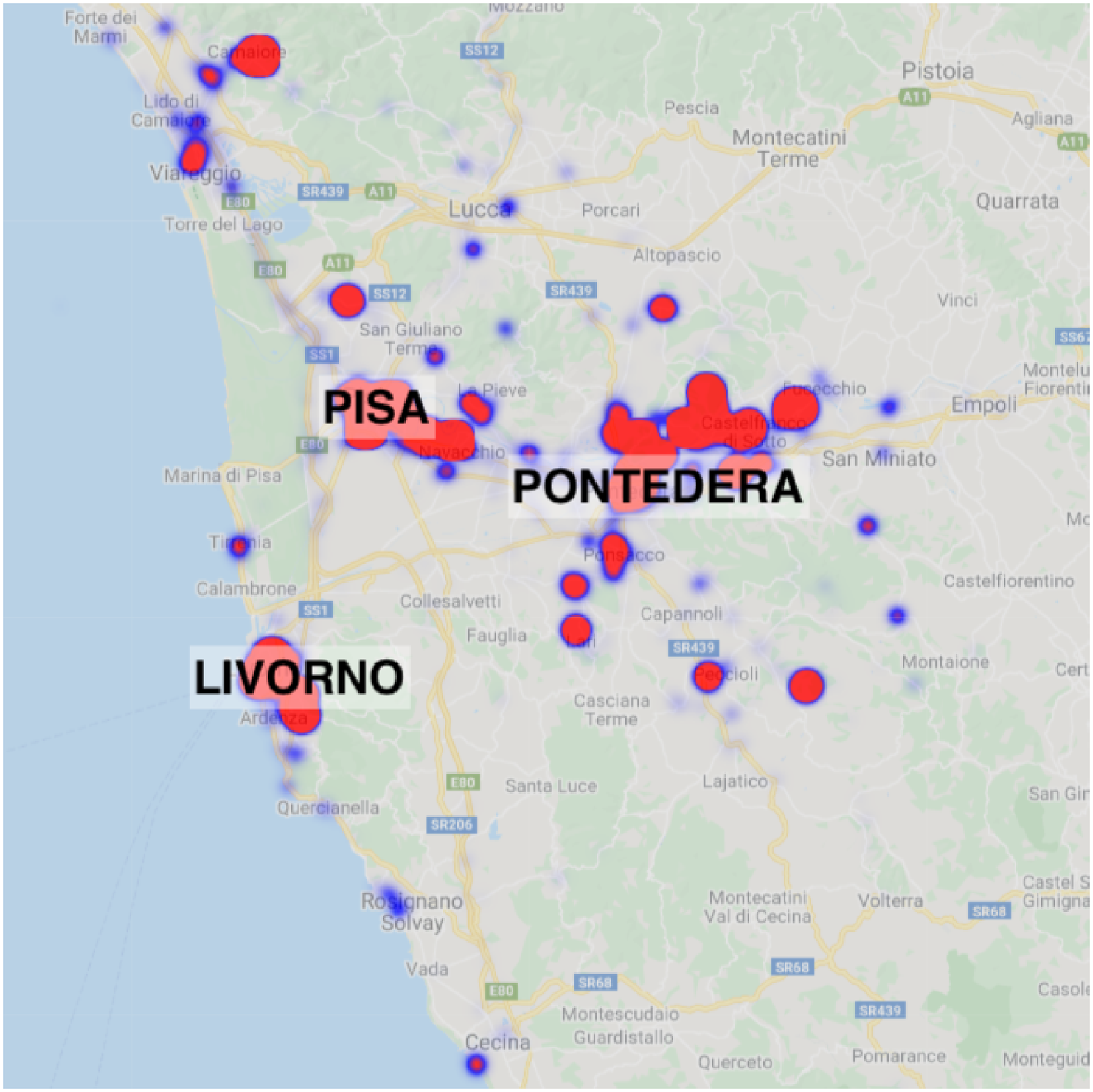}
        \caption{}
        \label{fig:loc_exp_area}
    \end{subfigure}

    \caption{Heatmap of the user's locations.}
    \label{fig:map_locations}
\end{figure}

%% file: fig_da_labels.tex
\begin{figure}[t]
    \centering
    \begin{subfigure}{\textwidth}
        \centering
        \includegraphics[width=\linewidth]{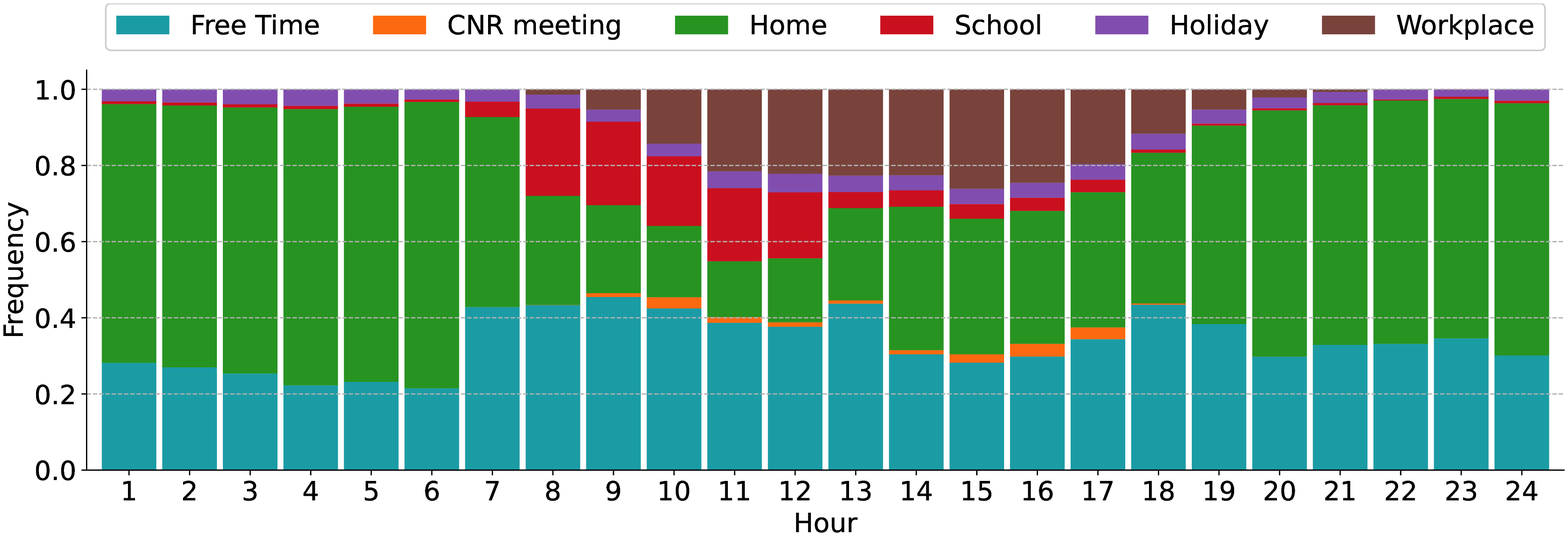}
        \caption{}
        \label{fig:labels_weekdays}
    \end{subfigure}\\
    \begin{subfigure}{\textwidth}
        \centering
        \includegraphics[width=\linewidth]{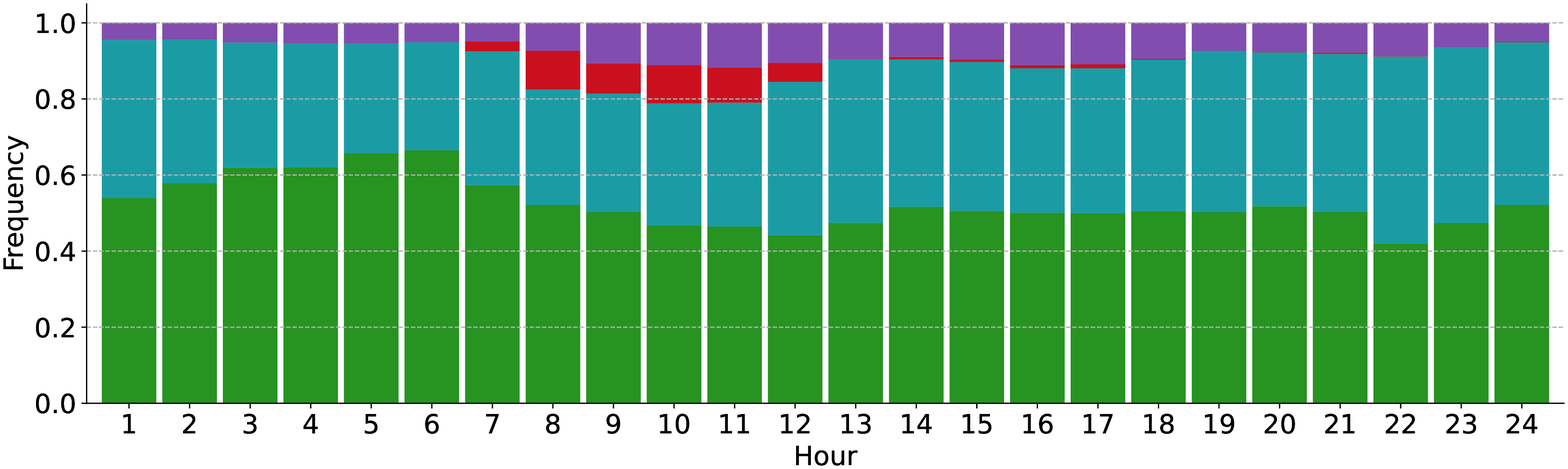}
        \caption{}
        \label{fig:labels_weekend}
    \end{subfigure}
    \caption{Labels distribution per hour during (a) weekdays and (b) the weekend. Best viewed in color.}
    \label{fig:da_labels_distribution}
\end{figure}

%% file: fig_data_analysis_osn.tex
\begin{figure}[t]
    \centering
    \begin{subfigure}{.49\textwidth}
        \centering
        \includegraphics[width=.95\linewidth]{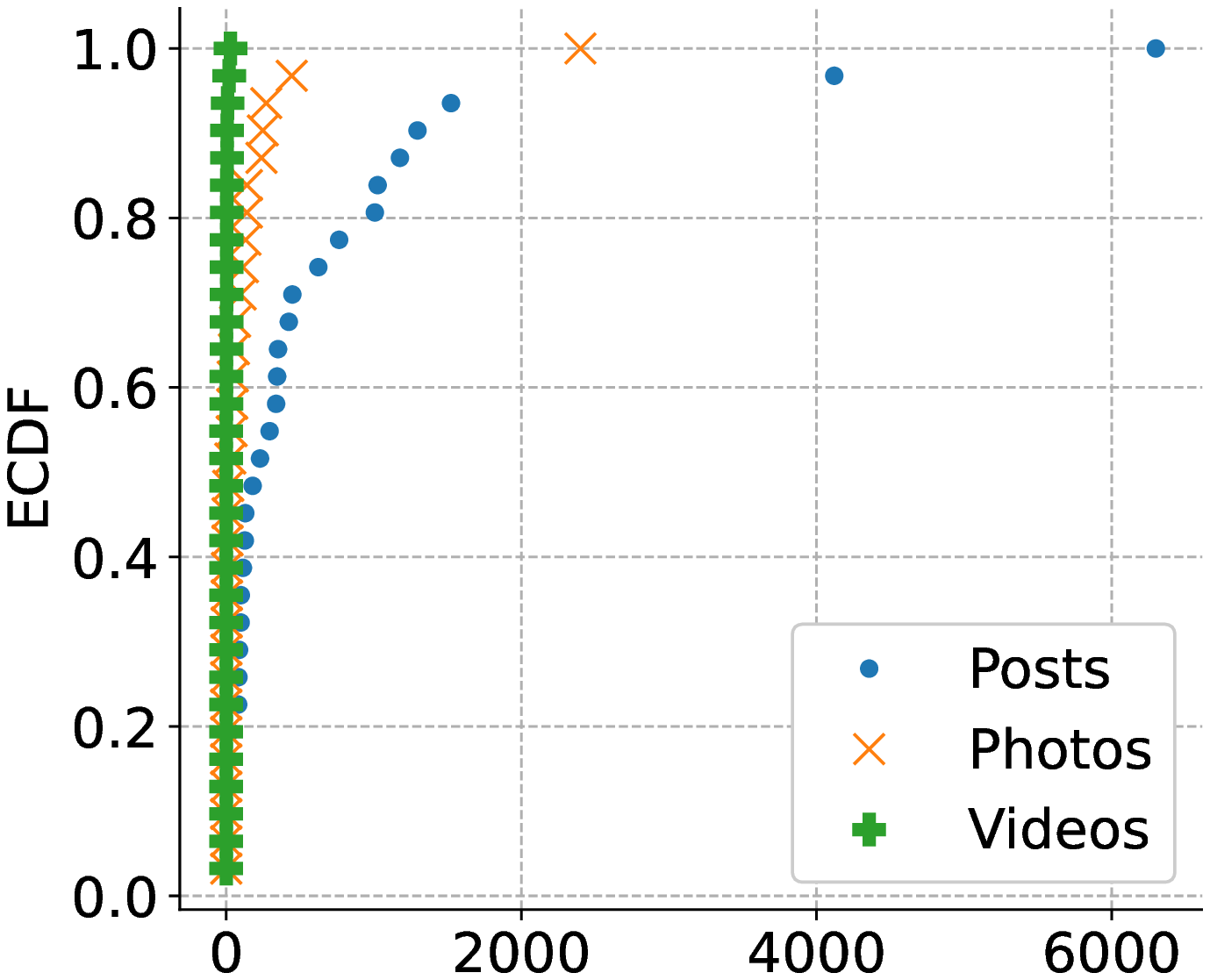}
        \caption{}
        \label{fig:osn_shared_items}
    \end{subfigure}
    \begin{subfigure}{.49\textwidth}
        \centering
        \includegraphics[width=.95\linewidth]{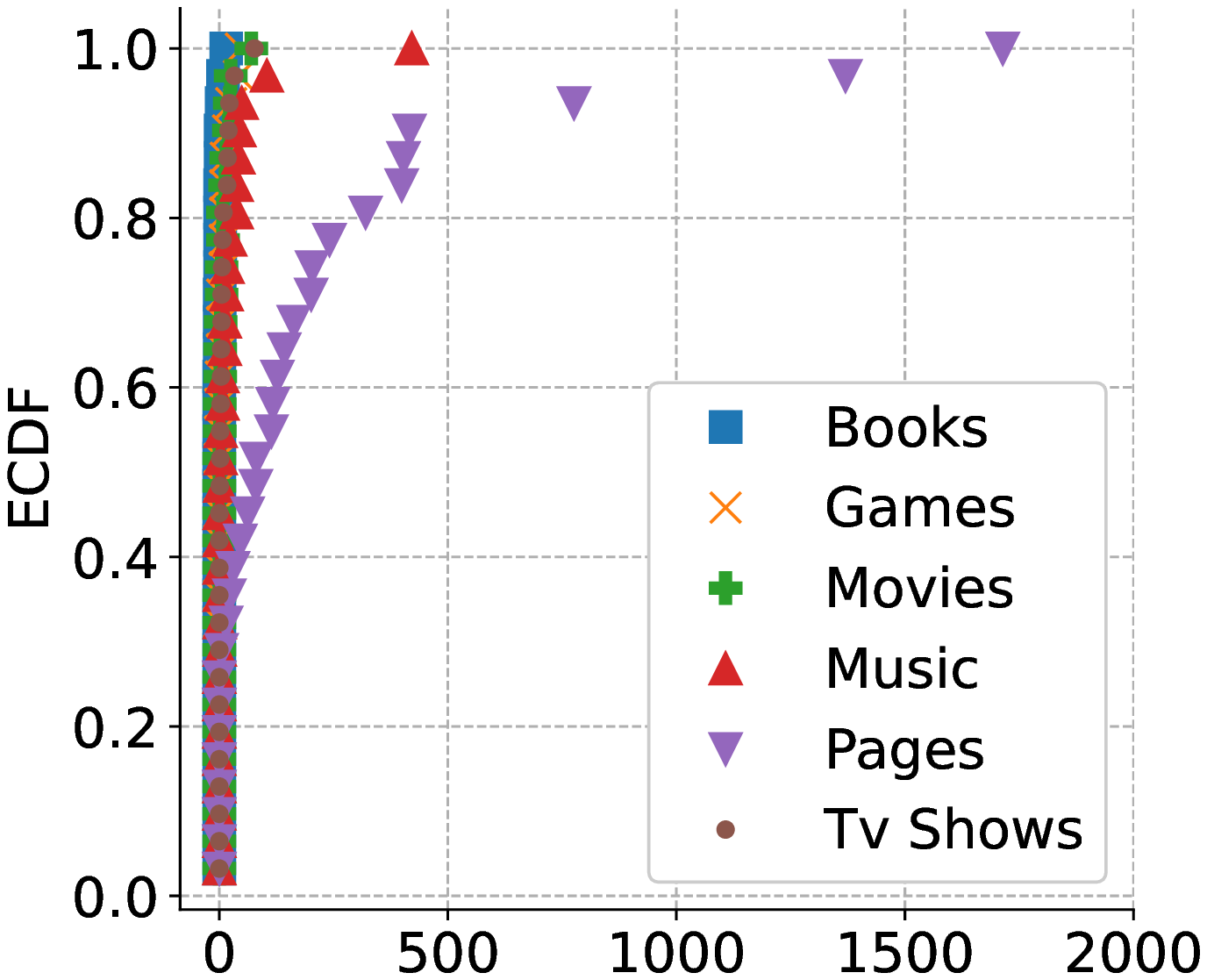}
        \caption{}
        \label{fig:osn_liked_items}
    \end{subfigure}
    \caption{Empirical Cumulative Distribution (ECDF) of (a) contents shared by the users on Facebook, and (b) the type of public pages they follow.}
    \label{fig:osn_stats}
\end{figure}

%% file: social_lp_parameters_tuning.tex
\begin{table}[t]
\centering
\footnotesize
\caption{Hyperparameters tuned for each method and their best values (in boldface).}
\begin{tabular}{lll}
\toprule
\textbf{Algorithm} & \textbf{Parameter} & \textbf{Values} \\\midrule
\multirow{4}{*}{DW}  & dimensions         & [\textbf{32}, 64, 128]                          \\
\multicolumn{1}{c}{} & walk-length        & [10, 50, \textbf{100}]                          \\
\multicolumn{1}{c}{} & number-walks       & [\textbf{10}, 50, 100]                          \\
\multicolumn{1}{c}{} & window-size        & [5, \textbf{10}, 20]                            \\
\midrule
\multirow{6}{*}{N2V} & dimensions         & [\textbf{32 (p)}, 64, \textbf{128 (c)}]         \\
                     & walk-length        & [10, \textbf{50 (p)}, \textbf{100 (c)}]         \\
                     & number-walks       & [10, \textbf{50 (p)}, \textbf{100 (c)}]         \\
                     & window-size        & [5, 10, \textbf{20 (p/c)}]                      \\
                     & p                  & [\textbf{1 (p/c)}, 10, 50]                      \\
                     & q                  & [\textbf{1 (c)}, 10, \textbf{50 (p)}]           \\
\midrule
\multirow{5}{*}{ANN} & layers             & [\textbf{1}, 2, 3]                              \\
                     & hidden-units       & [5, 10, 30, \textbf{50}, 100]                   \\
                     & learning-rate      & [0.001, 0.01, \textbf{0.1}]                     \\
                     & dropout-rate       & $[\textbf{0}, 0.1, 0.2, 0.3, \dots, 0.8, 0.9]$      \\
                     & batch-size         & [\textbf{32}, 64, 128]                          \\
                     & epochs             & [200, 500, \textbf{1000}]                       \\
\bottomrule
\end{tabular}
\label{tab:social_link_prediction_tuning}
\end{table}

%% file: social_lp_results.tex
\begin{table}[t]
\caption{Social link prediction results by using the physical and combined social information with the following algorithms: Common Neighbors (CN), Resource Allocation Index (RAI), Jaccard Index (JI), Adamic Adar Index (AAI), Preferential Attachment (PA), Node Embedding Heuristic (NEH), DeepWalk (DW), Node2Vec (N2V), and Neural Network (NN).}
\begin{tabularx}{\textwidth}{lrrrrrrrrr}
    \toprule
    \textbf{Data} & \textbf{CN} & \textbf{RAI} & \textbf{JI} & \textbf{AAI} & \textbf{PA} & \textbf{NEH} & \textbf{DW} & \textbf{N2V} & \textbf{NN} \\
    \midrule
    Physical & 0.88 & 0.89 & 0.88 & 0.89 & 0.67 & 0.83 & 0.81 & 0.84 & -             \\
    Combined &  -   &  -   &   -  &  -   &   -  &   -  &  -   & 0.88 & \textbf{0.96} \\
    \bottomrule
\end{tabularx}
\label{tab:social_link_prediction_results}
\end{table}

%% file: context_recognition_parameters.tex
\begin{table}[t]
\centering
\footnotesize
\caption{Grid search performed for the context recognition system. The best hyperparameters are highlighted in boldface.}
\begin{tabular}{lll}
\toprule
\textbf{Algorithm} & \textbf{Parameter} & \textbf{Values} \\\midrule
\multirow{3}{*}{DT}  & max-depth          & $[1, \dots, \mathbf{15}, \dots, 50]$            \\
\multicolumn{1}{c}{} & min-samples-leaf   & $[2,\dots, \mathbf{45}, \dots, 50]$             \\
\multicolumn{1}{c}{} & min-samples-split  & $[5, 10, 20, 50, \mathbf{100}]$                 \\
\midrule
\multirow{6}{*}{ANN} & layers             & [1, 2, \textbf{3}]                              \\
                     & hidden-units       & [100, \textbf{200}]                             \\
                     & learning-rate      & [\textbf{0.001}, 0.01, 0.1]                     \\
                     & dropout-rate       & $[0, 0.1, \textbf{0.2}, 0.3, \dots, 0.8, 0.9]$      \\
                     & batch-size         & [\textbf{128}, 512]                             \\
                     & epochs             & [50, \textbf{100}, 200]                         \\
\bottomrule
\end{tabular}
\label{tab:context_recognition_parameters}
\end{table}

%% file: fig_context_recognition_results.tex
\begin{figure}[t]
    \centering
    \begin{subfigure}{.49\textwidth}
        \centering
        \includegraphics[width=.95\linewidth]{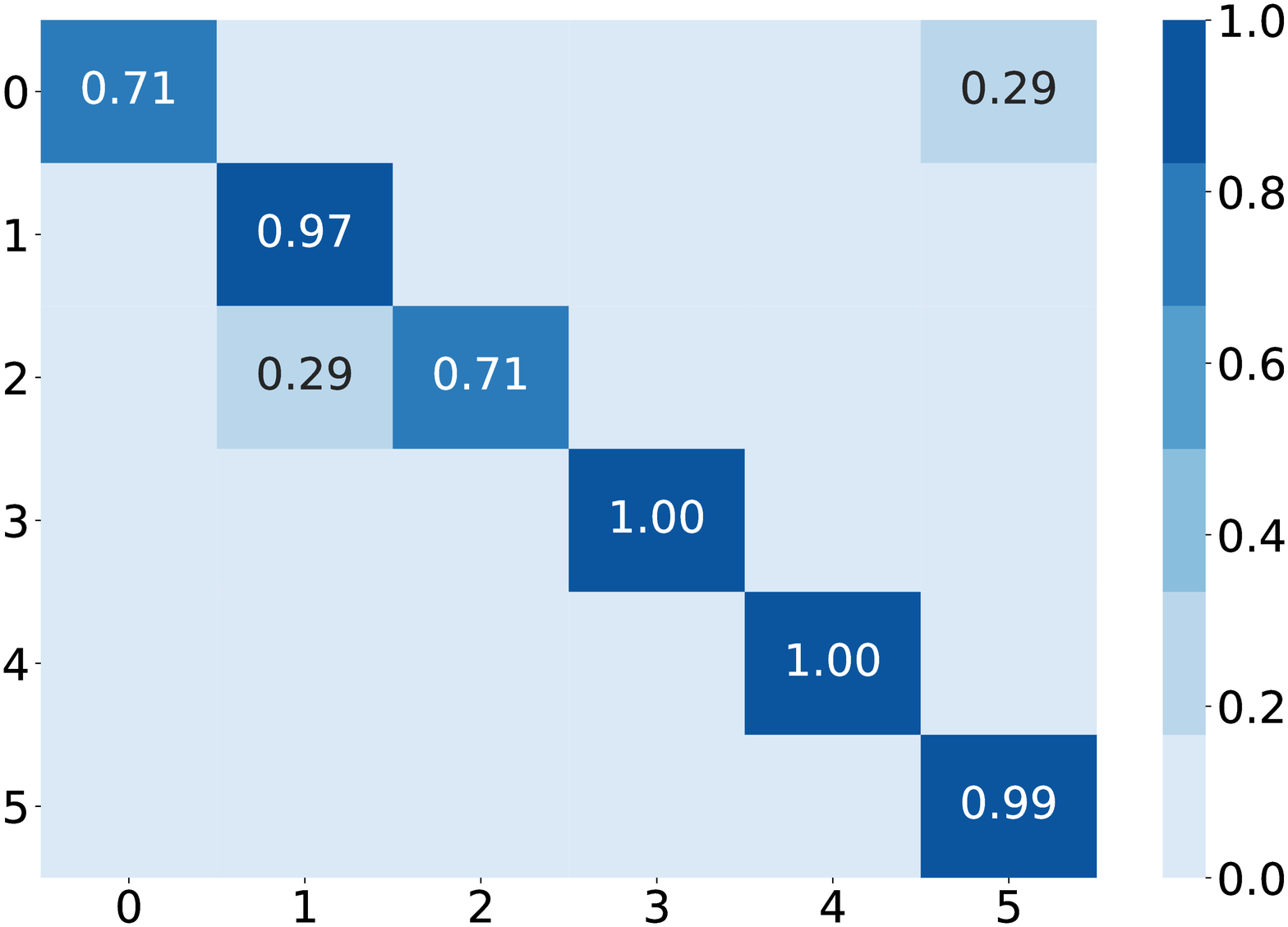}
        \caption{}
        \label{fig:cr_DT_results}
    \end{subfigure}
    \begin{subfigure}{.49\textwidth}
        \centering
        \includegraphics[width=.95\linewidth]{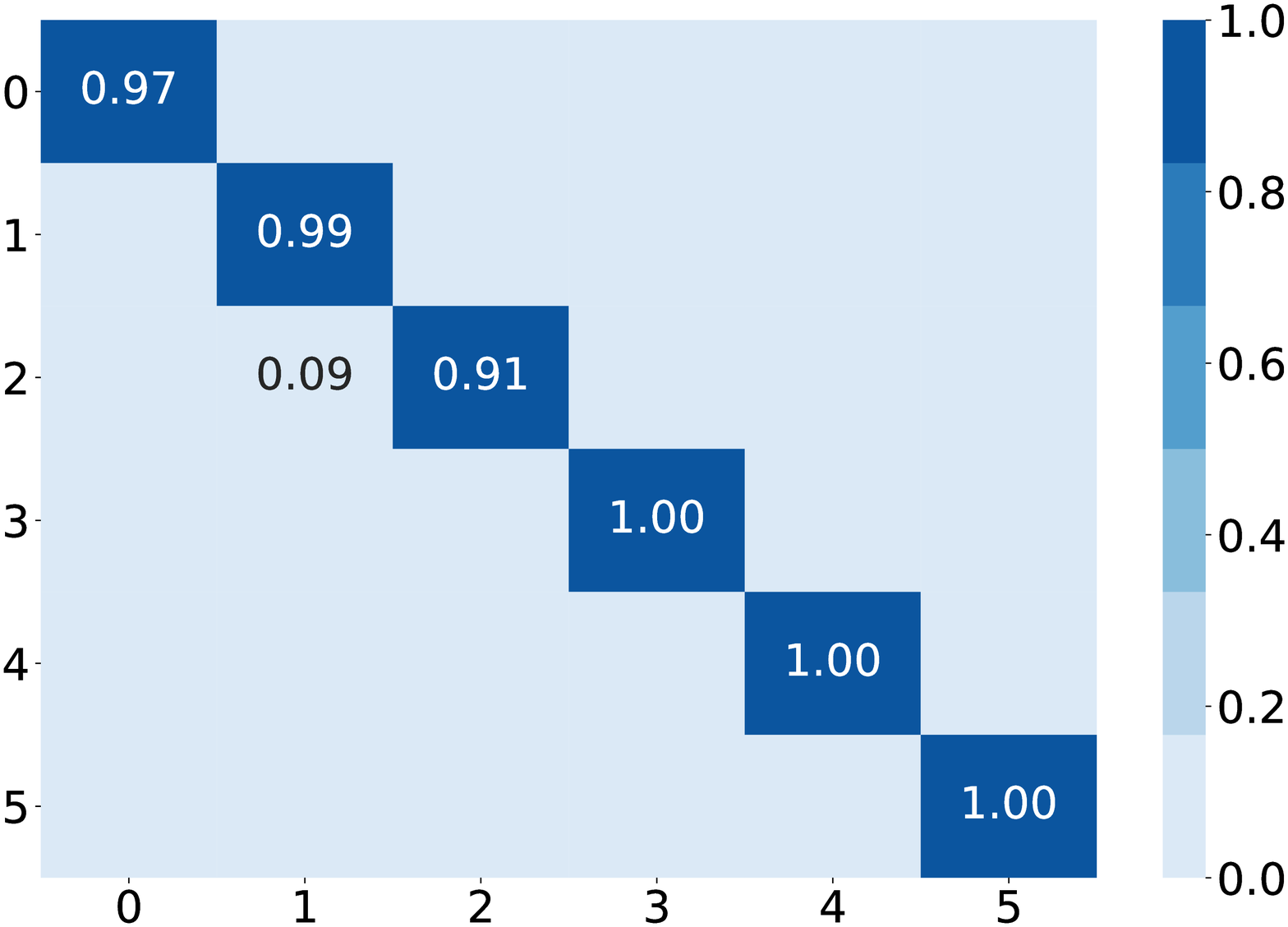}
        \caption{}
        \label{fig:cr_NN_results}
    \end{subfigure}
    \caption{Context recognition results. Confusion matrices of the predictions produced by (a) Decision Tree and (b) Deep Neural Network.}
    \label{fig:context_recognition_results}
\end{figure}

%% file: cars_parameters.tex
\begin{table}[t]
\centering
\footnotesize
\caption{Grid search performed for the P-CARS models. The best hyperparameters are highlighted in boldface.}
\begin{tabular}{lll}
\toprule
\textbf{Algorithm} & \textbf{Parameter} & \textbf{Values} \\\midrule
\multirow{3}{*}{DT}  & max-depth          & $[1, \dots, \mathbf{11}, \dots, 50]$            \\
\multicolumn{1}{c}{} & min-samples-leaf   & $[\mathbf{1}, \dots, 50]$             \\
\multicolumn{1}{c}{} & min-samples-split  & $[\mathbf{2}, 5, 10, 20, 50, 100]$                 \\
\midrule
\multirow{6}{*}{ANN} & layers             & [\textbf{1}, 2, 3]                              \\
                     & hidden-units       & [100, \textbf{200}]                             \\
                     & learning-rate      & [0.001, 0.01, \textbf{0.1}]                     \\
                     & dropout-rate       & $[0, 0.1, 0.2, 0.3, \dots, \mathbf{0.8}, 0.9]$  \\
                     & batch-size         & [128, 512, \textbf{1024}]                       \\
                     & epochs             & [50, 100, \textbf{200}]                         \\
\bottomrule
\end{tabular}
\label{tab:cars_parameters}
\end{table}

%% file: cars_results.tex
\begin{table}[t]
\centering
\footnotesize
\caption{Recommendations accuracy.}
\begin{tabular}{lrr}
    \toprule
    \textbf{Model} & \textbf{DT} & \textbf{NN}\\
    \midrule
    BASE & \textbf{0.963} & 0.942\\
    CARS &  0.984 & \textbf{0.994}\\
    \bottomrule
\end{tabular}
\label{tab:cars_results}
\end{table}

%% file: fig_cars_nn.tex
\begin{figure}[t]
    \centering
    \begin{subfigure}{.48\textwidth}
        \centering
        \includegraphics[width=\linewidth]{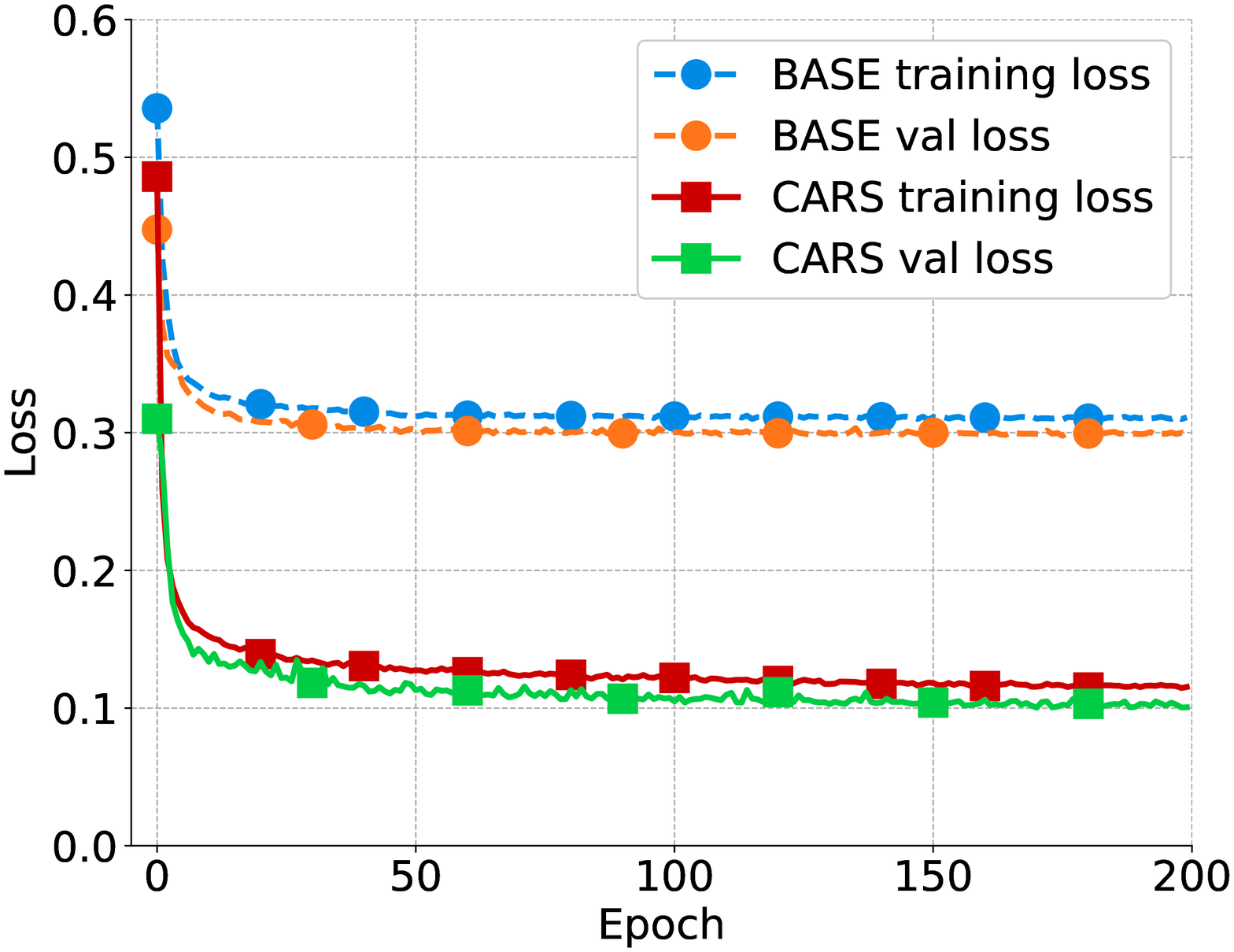}
        \caption{}
        \label{fig:cars_nn_loss}
    \end{subfigure}
    \begin{subfigure}{.48\textwidth}
        \centering
        \includegraphics[width=\linewidth]{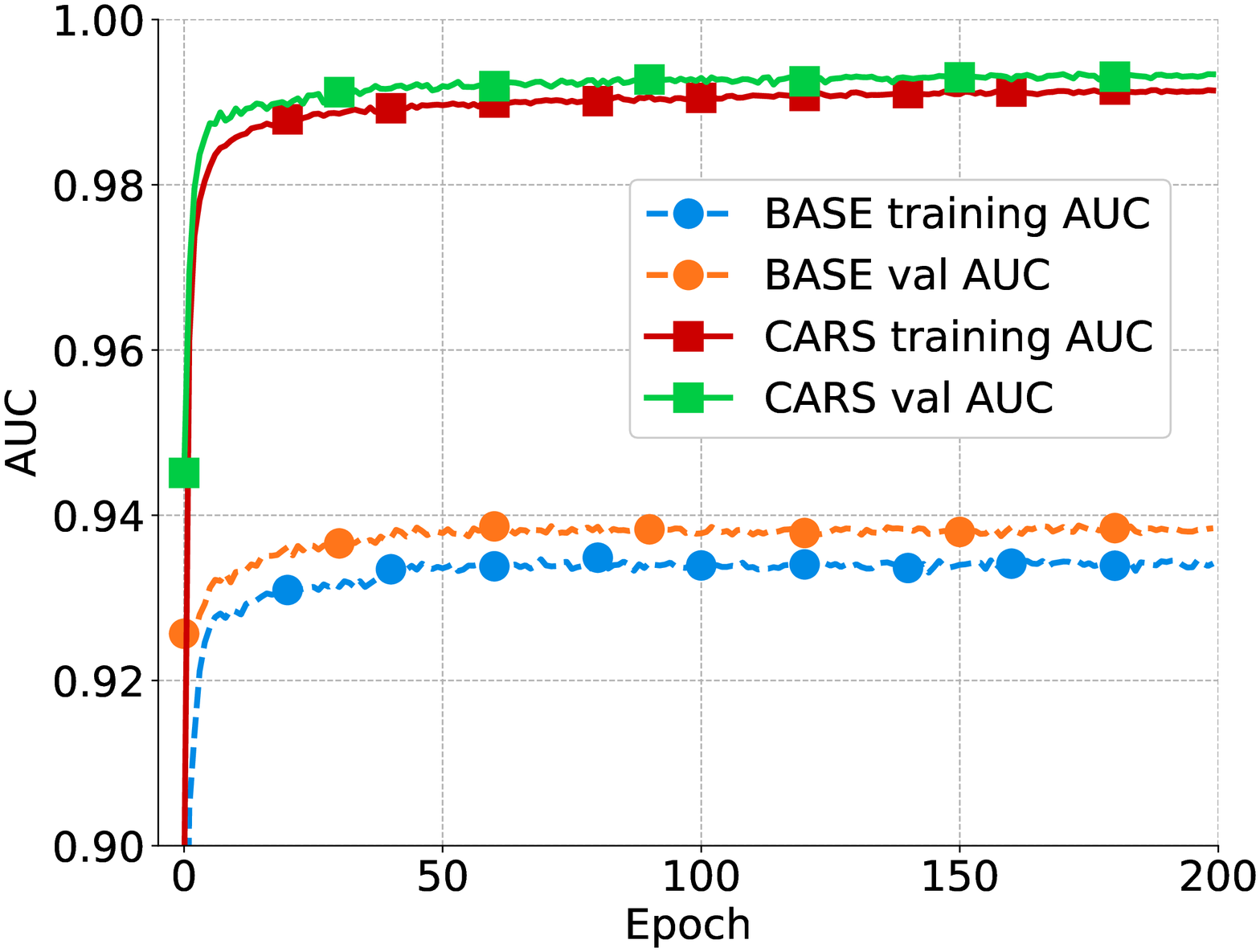}
        \caption{}
        \label{fig:cars_nn_auc}
    \end{subfigure}
    \caption{Training and validation loss (a) and AUC (b) performances obtained by the neural network model by using just the user-item information (BASE) and by using the additional context data (CARS).}
    \label{fig:cars_nn}
\end{figure}